\documentclass[10pt]{article} % For LaTeX2e
\usepackage[accepted]{tmlr}
\input{math_commands.tex}

\usepackage{hyperref}
\usepackage{url}
\usepackage{tabularx}
\usepackage{multirow, multicol}
\usepackage{graphicx}
\usepackage{booktabs}
\usepackage{xspace}
\usepackage{xcolor}
\definecolor{lightgray}{gray}{0.9}
\usepackage{colortbl}

\title{Part-aware Prompted Segment Anything Model \\ for Adaptive Segmentation}

\author{\name Chenhui Zhao \email chuizhao@umich.edu \\
\addr Department of Computer Science and Engineering, University of Michigan
\vspace{-5pt}
\AND
\name Liyue Shen \email liyues@umich.edu \\
\addr Department of Electrical and Computer Engineering, University of Michigan
}

\def\month{05}  % Insert correct month for camera-ready version
\def\year{2025} % Insert correct year for camera-ready version
\def\openreview{\url{https://openreview.net/forum?id=cCQKwd5MFP&noteId=GEyhAYE7Q6}} 
\begin{document}
\maketitle
\vspace{-10pt}
\begin{abstract}
    Precision medicine, such as patient-adaptive treatments assisted by medical image analysis, poses new challenges for segmentation algorithms in adapting to new patients, due to the large variability across different patients and the limited availability of annotated data for each patient.
    In this work, we propose a data-efficient segmentation algorithm, namely \emph{\textbf{P}art-aware \textbf{P}rompted \textbf{S}egment \textbf{A}nything \textbf{M}odel} ($\mathbf{{P}^{2}SAM}$).
    Without any model fine-tuning, $\text{P}^2\text{SAM}$ enables seamless adaptation to any new patients relying only on one-shot patient-specific data. 
    We introduce a novel part-aware prompt mechanism to selects multiple-point prompts based on the part-level features of the one-shot data, which can be extensively integrated into different promptable segmentation models, such as SAM and SAM~2. 
    Moreover, to determine the optimal number of parts for each specific case, we propose a distribution-guided retrieval approach that further enhances the robustness of the part-aware prompt mechanism.
    % To further promote the robustness of the part-aware prompt mechanism, we propose a distribution-guided retrieval approach to determine the optimal number of part-level features for a specific case.
    $\text{P}^2\text{SAM}$ improves the performance by $\texttt{+} 8.0\%$ and $\texttt{+} 2.0\%$ mean Dice score for two different patient-adaptive segmentation applications, respectively.
    In addition, $\text{P}^2\text{SAM}$ also exhibits impressive generalizability in other adaptive segmentation tasks in the natural image domain, \eg, $\texttt{+} 6.4\%$ mIoU within personalized object segmentation task. 
    The code is available at \url{https://github.com/Zch0414/p2sam}
\end{abstract}

\vspace{-10pt}
\section{Introduction}
\label{sec: introduction}
Advances in modern precision medicine and healthcare have emphasized the importance of patient-adaptive treatment~\citep{hodson2016precision}. For instance, in radiation therapy, the patient undergoing multi-fraction treatment would benefit from longitudinal medical data analysis that helps timely adjust treatment planning~\citep{sonke2019adaptive}. 
To facilitate the treatment procedure, such analysis demands timely and accurate automatic segmentation of tumors and critical organs from medical images, which has underscored the role of computer vision approaches for medical image segmentation tasks~\citep{hugo2016data, jha2020kvasir}.
Despite the great progress made by previous works~\citep{ronneberger2015u, isensee2021nnu}, their focus remains on improving the segmentation accuracy within a standard paradigm: trained on a large number of annotated data and evaluated on the \emph{internal} validation set.
However, patient-adaptive treatment presents unique challenges in adapting segmentation models to new patients: (1) the large variability across patients hinders direct model transfer, and (2) the limited availability of annotated training data for each patient prevents fine-tuning the model on a per-patient basis~\citep{chen2023patient}.
Overcoming these obstacles requires a segmentation approach that can reliably adapt to \emph{external} patients, in a data-efficient manner.

\begin{figure*}[t]
    \noindent
    \begin{minipage}[t]{.399\textwidth}
        \centering
        \hspace{-5pt}\includegraphics[width=\linewidth]{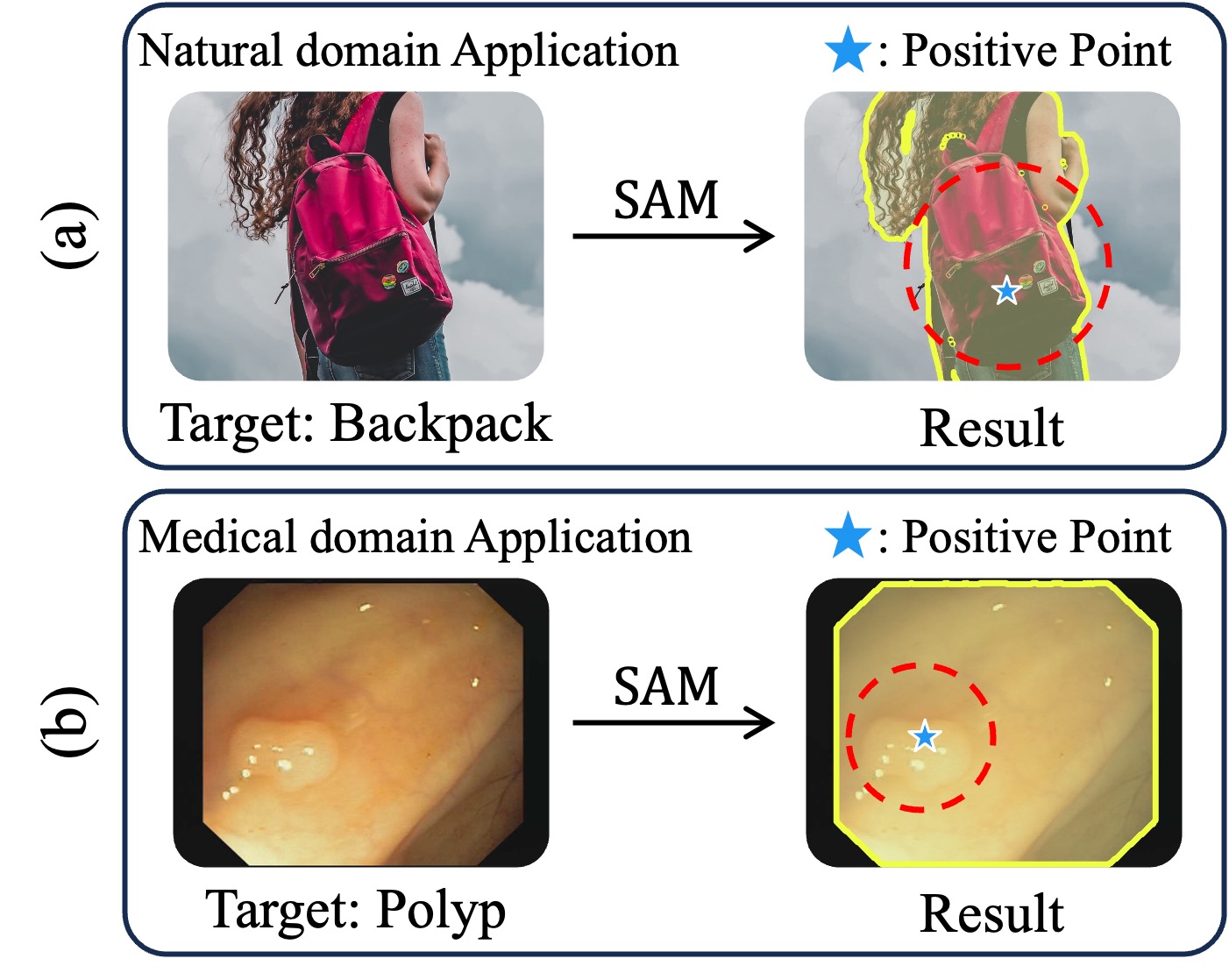} 
        \vspace{-10pt}
        \caption{Illustration of SAM's ambiguity property. 
        The ground truth is circled by a red dashed circle; the predicted mask is depicted by a yellow solid line.} 
        \label{fig: intro1}
    \end{minipage}%
    \hfill
    \begin{minipage}[t]{.561\textwidth} 
        \centering
        \hspace{-5pt}\includegraphics[width=\linewidth]{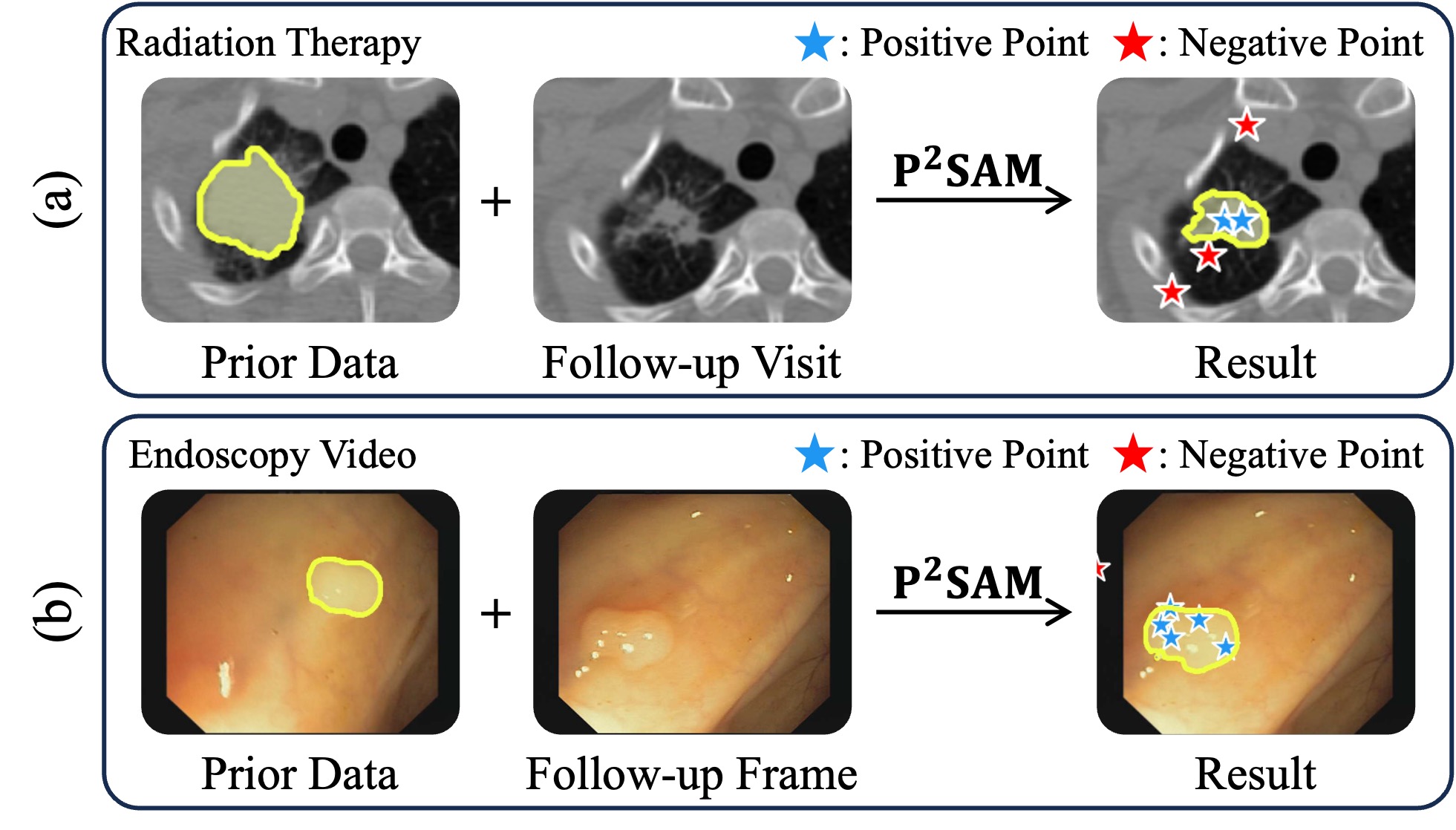}
        \vspace{-10pt}
        \caption{Illustration of two patient-adaptive segmentation tasks.
        \ppsam can segment the follow-up data by utilizing one-shot prior data as multiple-point prompts.
        Prior and predicted masks are depicted by a solid yellow line.
        }
        \label{fig: intro2}
    \end{minipage}%
    \vspace{-10pt}
\end{figure*}

In this work, we address the unmet needs of the patient-adaptive segmentation by formulating it as an in-context segmentation problem, where the \emph{context} is the prior data from a specific patient.
Such data can be obtained in a standard clinical protocol~\citep{chen2023patient}, therefore will not burden clinician. 
To this end, we propose $\mathbf{{P}^{2}SAM}$: \emph{\textbf{P}art-aware \textbf{P}rompted \textbf{S}egment \textbf{A}nything \textbf{M}odel}. 
Leveraging the promptable segmentation mechanism inherent in Segment Anything Model~(SAM)~\citep{kirillov2023segment}, our method seamlessly adapts to any \emph{external} patients relying only on one-shot patient-specific prior data without requiring additional training, thus in a data-efficient manner.
Beyond patient-adaptive segmentation, \ppsam also demonstrates strong generalizability in other adaptive segmentation tasks in the natural image domain, such as personalized segmentation~\citep{zhang2023personalize} and one-shot segmentation~\citep{liu2023matcher}.

In the original prompt mechanism of SAM~\citep{kirillov2023segment}, as illustrated in \Figref{fig: intro1}, a single-point prompt may result in ambiguous prediction, indicating the limitation in both natural domain and medical domain applications~\citep{zhang2023personalize, huang2024segment}.
To alleviate the ambiguity, following the statement in SAM, ``\emph{ambiguity is much rarer with multiple prompts}'', we propose a novel part-aware prompt mechanism that meticulously presents the prior data as multiple-point prompts based on part-level features.
As illustrated in \Figref{fig: intro2}, our method enables reliable adaptation to an \emph{external} patient across various tasks with one-shot patient-specific prior data.
To extract part-level features, we cluster the prior data into multiple parts in the feature space and computing the mean for each part.
Then, we select multiple-point prompts based on the cosine similarity between these part-level features and the follow-up data. 
The proposed approach can be generalized to different promptable segmentation models that support the point modality, such as SAM and its successor, SAM~2~\citep{ravi2024sam}. Here, we primarily utilize SAM as the backbone model, and SAM~2 will be integrated within the specific setting.

On the other hand, when the number of parts is set suboptimally, either more or less, the chance of encountering outlier prompts may increase. 
In the extreme, assigning all image patches to a single part produces an ambiguity-aware prompt~\citep{zhang2023personalize}, whereas assigning each image patch to a different part yields many outlier prompts~\citep{liu2023matcher}.
Determining the optimal number of parts is non-trivial, as it may vary across different cases.
Here, we introduced a novel distribution-guided retrieval approach to investigate the optimal number of parts required by each case. 
This retrieval approach is based on the distribution distance between the foreground feature of the prior image and the resulting feature obtained under the current part count. This principle is motivated by the fact that tumors and normal organs always lead to distinct feature distributions within medical imaging technologies~\citep{garcia2019clinical}.

With the aforementioned designs, \ppsam tackles a fundamental challenge—ambiguity—when adapting promptable segmentation models to specific applications. When ambiguity is not an issue, \ppsam enhances model generality by providing curated information.
The key contributions of this work lie in three-fold:
\begin{enumerate}
    \vspace{-10pt}
    \item 
    We formulate the patient-adaptive segmentation as an in-context segmentation problem, resulting in a data-efficient segmentation approach, \ppsam, that requires only one-shot prior data and no model fine-tuning. \ppsam functions as a generic segmentation algorithm, enabling efficient and flexible adaptation across different domains, tasks, and models.
    \vspace{-5pt}
    \item 
    We propose a novel part-aware prompt mechanism that can select multiple-point prompts based on part-level features. Additionally, we introduce a distribution-guided retrieval approach to determine the optimal number of part-level features required by different cases. 
    These designs significantly enhance the generalizability of promptable segmentation models.  
    \vspace{-5pt}
    \item 
    Our method largely benefits real-world applications like patient-adaptive segmentation, one-shot segmentation, and personalized segmentation. 
    Experiment results demonstrate that \ppsam improves the performance by $\texttt{+} 8.0\%$ and $\texttt{+} 2.0\%$ mean Dice score in two different patient-adaptive segmentation applications and achieves a new state-of-the-art result, \ie, $95.7\%$ mIoU on the personalized segmentation benchmark PerSeg.
\end{enumerate}

\section{Related Work}
\label{sec: related work}

\noindent\textbf{Segmentation Generalist.}
Over the past decade, various segmentation tasks including semantic segmentation~\citep{strudel2021segmenter, li2023paintseg}, instance segmentation~\citep{he2017mask, li2022hybrid}, panoptic segmentation~\citep{carion2020end, cheng2021per, li2022exploring}, and referring segmentation~\citep{li2023robust, zou2024segment} have been extensively explored for the image and video modalities. 
Motivated by the success of foundational language models~\citep{radford2018improving, radford2019language, brown2020language, touvron2023llama}, the computer vision research community is increasingly paying attention to developing more generalized models that can tackle various vision or multi-modal tasks, or called foundation models~\citep{li2022exploring, oquab2023dinov2, yan2023universal, wang2023images, wang2023seggpt, kirillov2023segment}. 
Notably, Segment Anything model~(SAM)~\citep{kirillov2023segment} and its successor, SAM~2~\citep{ravi2024sam} introduces a promptable model architecture, including the positive- and negative-point prompt; the box prompt; and the mask prompt. SAM and SAM~2 emerge with an impressive zero-shot interactive segmentation capability after pre-training on the large-scale dataset. The detail of SAM can be found in Appendix~\ref{appendix: sam overview}.

\noindent\textbf{Medical Segmentation.}
Given the remarkable generality of SAM and SAM~2, researchers within the medical image domain have been seeking to build foundational models for medical image segmentation~\citep{wu2023medical, wong2023scribbleprompt, wu2024one, zhang2024unleashing} in the same interactive fashion.
To date, ScribblePrompt~\citep{wong2023scribbleprompt} and One-Prompt~\citep{wu2024one} introduce a new prompt modality—scribble—that provides a more flexible option for clinician usage.
MedSAM~\citep{ma2024segment} fine-tunes SAM on an extensive medical dataset, demonstrating significant performance across various medical image segmentation tasks. Its successor~\citep{ma2024segment2} incorporates SAM~2 to segment a 3D medical image volume as a video.
However, these methods rely on clinician-provided prompts for promising segmentation performance. Moreover, whether these methods can achieve zero-shot performance as impressive as SAM and SAM 2 remains an open question that requires further investigation~\citep{ma2024segment2}.

\noindent\textbf{In-Context Segmentation.}
The concept of in-context learning is first introduced as a new paradigm in natural language processing~\citep{brown2020language}, allowing the model to adapt to unseen input patterns with a few prompts and examples, without the need to fine-tune the model. 
Similar ideas~\citep{rakelly2018conditional, sonke2019adaptive, li2023robust} have been explored in segmentation tasks. For example, few-shot segmentation~\citep{rakelly2018conditional, wang2019panet, liu2020part, leng2024self} like PANet~\citep{wang2019panet}, aims to segment new classes with only a few examples; in adaptive therapy~\citep{sonke2019adaptive}, several works~\citep{elmahdy2020patient, wang2020predicting, chen2023patient} attempt to adapt a segmentation model to new patients with limited patient-specific data, but these methods require model fine-tuning in different manners. 
Recent advancements, such as Painter~\citep{wang2023images} and SegGPT~\citep{wang2023seggpt} pioneer novel in-context segmentation approaches, enabling the timely segmentation of images based on specified image-mask prompts. SEEM~\citep{zou2024segment} further explores this concept by investigating different prompt modalities.
More recently, PerSAM~\citep{zhang2023personalize} and Matcher~\citep{liu2023matcher} have utilized SAM to tackle few-shot segmentation through the in-context learning fashion. 
% PerSAM introduces a novel task, known as personalized object segmentation, which aims at adapting SAM to new views of a specific object. 
However, PerSAM prompts SAM with a single point prompt, causing ambiguity in segmentation results and therefore requires an additional fine-tuning strategy.
% On the other hand, Matcher enhances segmentation accuracy by sampling multiple sets of point prompts. However, Matcher's prompt generation mechanism is based on patch-level features. 
Matcher samples multiple sets of point prompts but based on patch-level features. This mechanism makes Matcher dependent on DINOv2~\citep{oquab2023dinov2} to generate prompts, which is particularly pre-trained under a patch-level objective. Despite this, Matcher still generates a lot of outlier prompts, therefore relies on a complicated framework to filter the outlier results.
% Thus, Matcher relies on a complicated framework and lacks flexibility and robustness when integrated into other vision backbones, including SAM.

In this work, we address the patient-adaptive segmentation problem, also leveraging SAM's promptable ability. Our prompt mechanism is based on part-level features, which will not cause ambiguity and are more robust than patch-level features. The optimal number of parts for each case is determined by a distribution-guided retrieval approach, further enhancing the generality of the part-aware prompt mechanism.

\section{Method}
\label{sec: method}
In \Secref{subsec: problem setting}, we define the problem within the context of patient-adaptive segmentation. In \Secref{subsec: methodology overview}, we present the proposed methodology, \ppsam, within a broader setting of adaptive segmentation.
% Note that our method can adapt to various domains. Therefore, we incorporate natural image illustrations in this section to provide a more intuitive understanding.
In \Secref{subsec: adapt sam to medical image domain if needed}, we introduce an optional fine-tuning strategy when adapting the backbone model to medical image domain is required.

\subsection{Problem Setting}
\label{subsec: problem setting}

Our method aims to adapt a promptable segmentation model to \emph{external} patients, with only one-shot patient-specific prior data. As shown in \Figref{fig: intro2}, such data can be obtained in a standard clinical protocol, either from the initial visit of radiation therapy or the first frame of medical video. The prior data includes a reference image $I_R$ and a mask $M_R$ delineating the segmented object. 
Given a target image, $I_T$, our goal is to predict its mask $M_T$, 
without additional human annotation costs or model training burdens.

\subsection{Methodology Overview}
\label{subsec: methodology overview}

\begin{figure*}[t]
  \centering
  \includegraphics[width=0.96\linewidth]{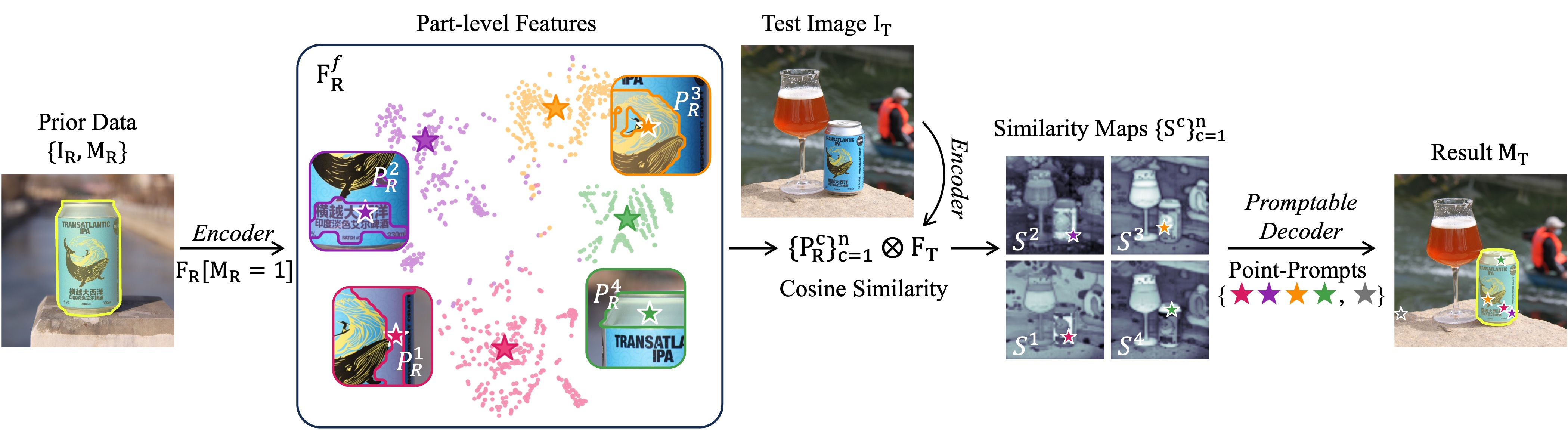}
  \vspace{-10pt}
  \caption{Illustration of the part-aware prompt mechanism. Masks are depicted by a yellow solid line. We first cluster foreground features in the reference image into part-level features. Then, we select multiple-point prompts based on the cosine similarity~($\otimes$ in the figure) between these part-level features and target image features. A colorful star, matching the color of the corresponding part, denotes a positive-point prompt, while a gray star denotes a negative-point prompt. These prompts are subsequently fed into the promptable decoder to do prediction.}
  \label{fig: method1}
  \vspace{-10pt}
\end{figure*}

The setting described in \Secref{subsec: problem setting} can be extended to other adaptive segmentation tasks in the natural image domain where the target image represents a new view or instance of the object depicted in the prior data. As shown in \Figref{fig: method1}, we illustrate our part-aware prompt mechanism using a natural image to clarify the significance of each part. Additional visualizations for parts in medical images are provided in Appendix~\ref{appendix: additional visualizations}. Since no part-level definitions exist for the two diseases studied in this work, we refer these parts as data-driven parts.

\noindent \textbf{Part-aware Prompt Mechanism.} 
We utilize SAM~\citep{kirillov2023segment} as the backbone model here, but our approach can be generalized to other promptable segmentation models that support the point prompt modality, such as SAM~2~\citep{ravi2024sam}.
Given the reference image-mask pair from the prior data, $\{I_R, M_R\}$, \ppsam first apply SAM's \emph{Encoder} to extract the visual features $ F_R \in \mathbb{R}^{h \times w \times d} $ from the reference image $ I_R $. Then, we utilize the reference mask $M_R$ to select foreground features $F^{f}_{R}$~($F_R[M_R = 1]$) by:
\begin{equation}
    F^{f}_{R} = \{ {F_R}_{ij} \mid {M_R}_{ij} = 1, \forall (i,j)\in \mathcal{I}^{h \times w} \}
\label{equ: mask pooling}
\end{equation}
where $\mathcal{I}^{h \times w}$ is the spatial coordinate set of $F_R$.
We cluster $F^{f}_{R}$ with \textit{k}-mean\texttt{++}~\citep{arthur2007k} into $n$ parts.
Then, we obtain $n$ part-level features $\left\{ P^{c}_R \right\}^{n}_{c=1} \in \mathbb{R}^{n \times d}$ by computing the mean of each part. 
In \Figref{fig: method1}, we showcase an example of $n\texttt{=}4$. Each part-level feature $P^{c}_R$ is represented by a colorful star in the foreground feature space. We further align the features of each part with pixels in the RGB space, thereby contouring the corresponding regions for each part in the image, respectively. 
% We find that SAM's encoder tends to cluster features together based on texture features, such as the characters and pictures depicted on the can. 
We extract the features $F_T \in \mathbb{R}^{h \times w \times d}$ from the target image $I_T$ using the same \emph{Encoder}, and compute similarity maps $\left\{ S^{c} \right\}^{n}_{c=1} \in \mathbb{R}^{n \times h \times w}$ based on the cosine similarity between part-level features $\left\{ P^{c}_R \right\}^{n}_{c=1}$ and $F_T$ by:
\begin{equation}
    {S^{c}}_{ij} = \frac{P^{c}_{R} \cdot {F_T}_{ij}}{{\left \| P^{c}_{R} \right \|}_{2} \cdot {\left \| {F_{T}}_{ij} \right \|}_{2}}
\label{equ: cosine similarity}
\end{equation}
We determine $n$ positive-point prompts $ \left\{\mathit{Pos}^{c}\right\}_{c=1}^{n} $ with the highest similarity score on each similarity map $S^{c}$. In \Figref{fig: method1}, each prompt $Pos^{c}$ is depicted as a colorful star on the corresponding similarity map $S^{c}$. 

For natural images, the background of the reference image and the target image may exhibit little correlation. Thus, following the approach in PerSAM~\citep{zhang2023personalize}, we choose one negative-point prompt $\left\{\mathit{Neg}\right\}$ with the lowest score on the average similarity map $\frac{1}{n}\sum^{n}_{c=1}S^{c}$. $\left\{\mathit{Neg}\right\}$ is depicted as the gray star in \Figref{fig: method1}.
However, for medical images, the background of the reference image is highly correlated with the background of the target image, usually both representing normal anatomical structures. As a result, in medical images, shown as \Figref{fig: intro2} in \Secref{sec: introduction}, we identify multiple negative-point prompts $ \left\{\mathit{Neg}^{c}\right\}_{c=1}^{n} $ from the background. 
This procedure mirrors the selection of multiple positive-point prompts but we use background features $F^{b}_R$~($F_R[M_R=0]$). Finally, we send both positive- and negative-point prompts into SAM's \emph{Promptable Decoder} and get the predicted mask $M_T$ for the target image.

\begin{figure*}[t]
    \noindent
    \begin{minipage}[t]{.48\textwidth}
        \centering
        \includegraphics[width=\linewidth]{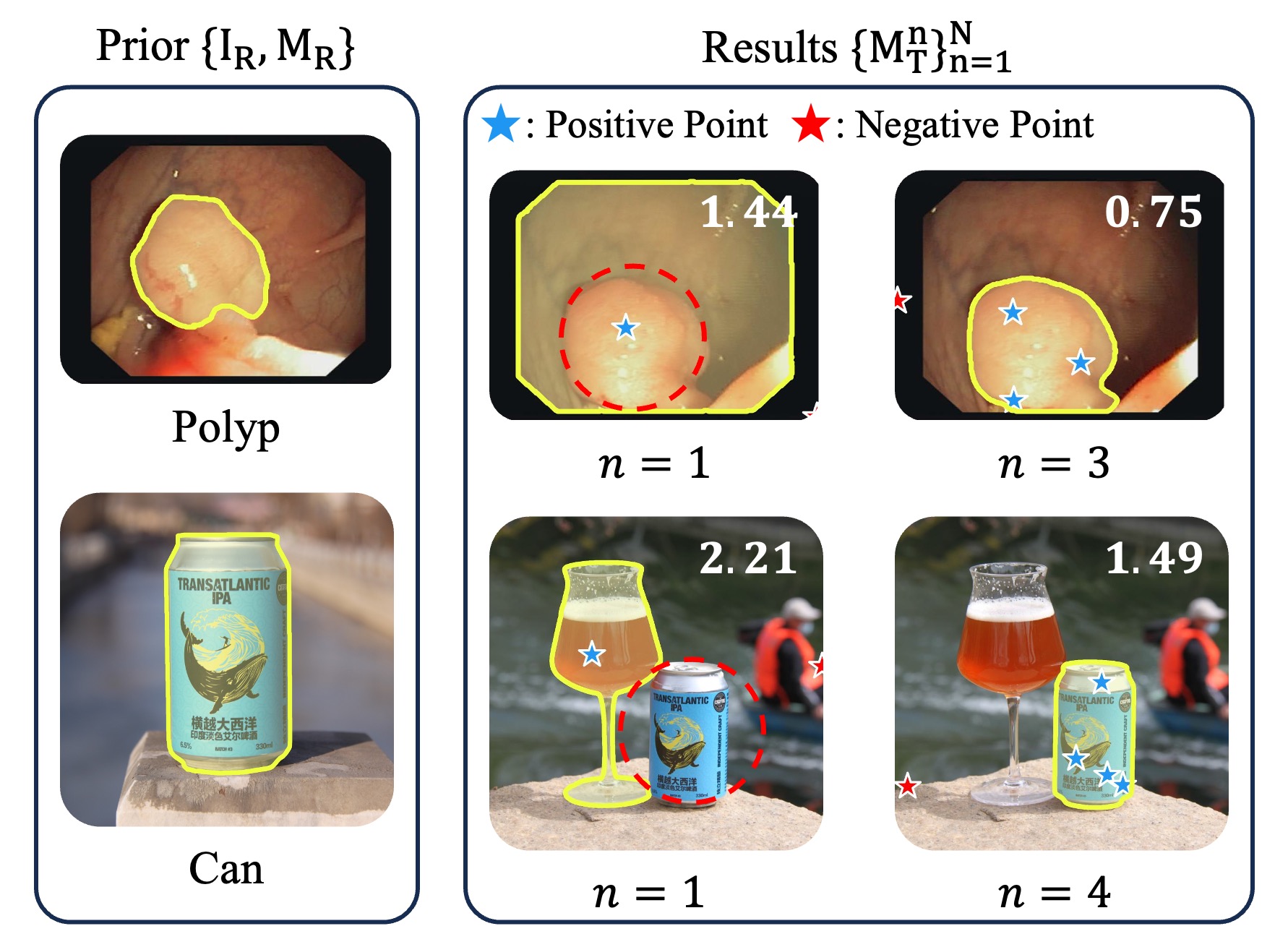} 
        \vspace{-25pt}
        \caption{Illustration of \ppsam's improvement. Wasserstein distances between the priors and results are shown in white.} 
        \label{fig: method2}
    \end{minipage}%
    \hfill
    \begin{minipage}[t]{.48\textwidth} 
        \centering
        \includegraphics[width=\linewidth]{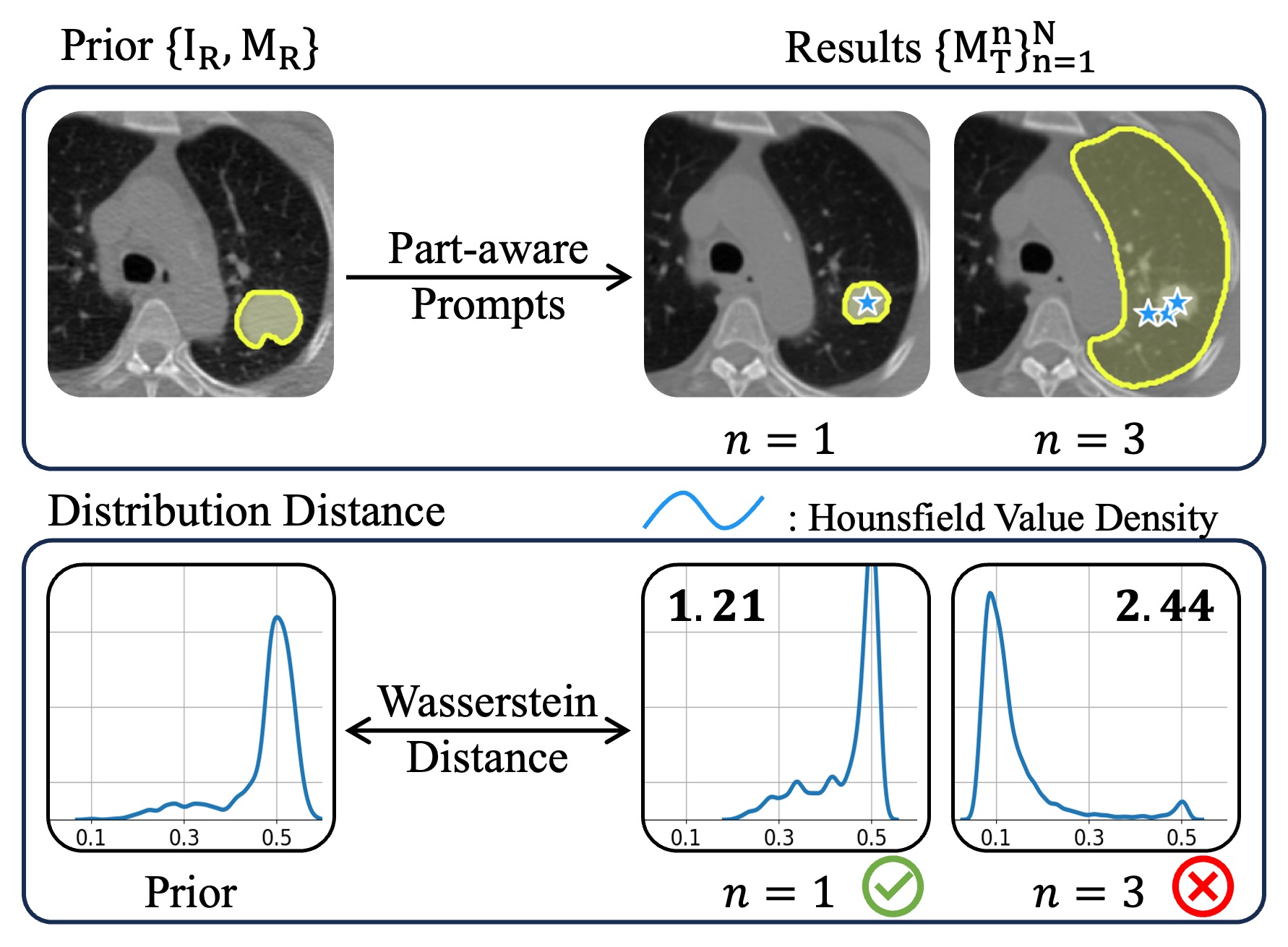}
        \vspace{-20pt}
        \caption{Illustration of the distribution-guided retrieval approach.}
        \label{fig: method3}
    \end{minipage}%
    \vspace{-5pt}
\end{figure*}

\noindent\textbf{Distribution-Guided Retrieval Approach.}
Improvements of the part-aware prompt mechanism are illustrated in \Figref{fig: method2}. The proposed approach can naturally avoid the ambiguous prediction introduced by SAM (\eg, polyp) and also improve precision (\eg, can). However, this approach may occasionally result in outliers, as observed in the segmentation example in \Figref{fig: method3}, $n\texttt{=}3$.
Therefore, we propose a distribution-guided retrieval approach to answer the question, ``\emph{How many part-level features should we choose for each case?}''. 
We assume the correct target foreground feature $F^{f}_{T}$~($F_{T}[M_{T}=1]$), and the reference foreground feature $F^{f}_{R}$ should belong to the same distribution. 
This assumption is grounded in the fact that tumors and normal organs will be reflected in distinct distributions by medical imaging technologies~\citep{garcia2019clinical}, also observed by the density of Hounsfield Unit value in \Figref{fig: method3}.
To retrieve the optimal number of parts for a specific case, we first define $N$ candidate part counts, and obtain $N$ part-aware candidate segmentation results $\{M^{n}_{T}\}^{N}_{n=1}$. 
After that, we extract $N$ sets of target foreground features $\{{F^{f(n)}_{T}}\}_{n=1}^{N}$. Following WGAN~\citep{arjovsky2017wasserstein}, we utilize Wasserstein distance $\mathcal{D}_{w}(\cdot, \cdot)$ to measure the distribution distance between reference foreground features $F^{f}_{R}$ and each set of target foreground features ${F^{f(n)}_{T}}$. We determine the optimal number of parts $n$ by:
\begin{equation}
n=\argmin_{n \in \left\{1, \cdots, N\right\}} \mathcal{D}_{w}(F^{f}_{R},{F^{f(n)}_{T}}),
\label{equ: distribution-guided retrieval}
\end{equation}
where the details of $\mathcal{D}_{w}(\cdot, \cdot)$ can be found in Appendix~\ref{appendix: equations}, Equation~\ref{equ: wasserstein distance}. The smaller distance value for the correct prediction in \Figref{fig: method2} indicates this approach can be extended to multiple image modalities.

\subsection{Adapt SAM to Medical Image Domain}
% if Needed}
\label{subsec: adapt sam to medical image domain if needed}

Segment Anything Model~(SAM)~\citep{kirillov2023segment} is initially pre-trained on the SA-1B dataset. Despite the large scale, a notable domain gap persists between natural and medical images. 
In more realistic medical scenarios, clinic researchers could have access to certain public datasets~\citep{aerts2015data, jha2020kvasir} tailored to specific applications, enabling them to fine-tune the model.
Nevertheless, even after fine-tuning, the model can still be limited to generalize across various \emph{external} patients from different institutions because of the large variability in patient population, demographics, imaging protocol, etc.
\ppsam can then be flexibly plugged into the fine-tuned model to enhance robustness on unseen patients.

Specifically, when demanded, we utilize \emph{internal} medical datasets~\citep{aerts2015data, jha2020kvasir} to fine-tune SAM. We try full fine-tune, and Low-Rank adaptation~(LoRA)~\citep{hu2021lora} for further efficiency. During the fine-tuning, similar to Med-SA~\citep{wu2023medical}, we adhere closely to the interactive training strategy outlined in SAM to maintain the interactive ability. Details can be found in Appendix~\ref{appendix: sam adaptation details}.
Then, we employ \emph{external} datasets~\citep{bernal2015wm, hugo2016data} obtained from various institutions to mimic new patient cases.
Note that there is no further fine-tuning on these datasets. 

\section{Experiments}
\label{sec: experiments}

In \Secref{subsec: experiment settings}, we introduce our experimental settings. In \Secref{subsec: quantitative results}, we evaluate the quantitative results of our approach. In \Secref{subsec: ablation study}, we conducted several ablation studies to investigate our designs. In \Secref{subsec: qualitative results}, we show qualitative results.

\subsection{Experiment Settings}
\label{subsec: experiment settings}

\noindent\textbf{Dataset.} 
We utilize a total of four medical datasets, including two \emph{internal} datasets: 
The NSCLC-Radiomics dataset~\citep{aerts2015data}, collected for non-small cell lung cancer~(NSCLC) segmentation, contains data from 422 patients. Each patient has a computed tomography~(CT) volume along with corresponding segmentation annotations.
The Kvasir-SEG dataset~\citep{jha2020kvasir}, contains $1000$ labeled endoscopy polyp images.
Two \emph{external} datasets from different institutions:
The 4D-Lung dataset~\citep{hugo2016data}, collected for longitudinal analysis, contains data from $20$ patients, within which $13$ patients underwent multiple visits, $3$ to $8$ visits for each patient. For each visit, a CT volume along with corresponding segmentation labels is available.
The CVC-ClinicDB dataset~\citep{bernal2015wm}, contains $612$ labeled polyp images selected from $29$ endoscopy videos.
During experiments, \emph{internal} datasets serve as the training dataset to adapt SAM to the medical domain, while \emph{external} datasets serve as unseen patient cases.

\noindent\textbf{Patient-Adaptive Segmentation Tasks.} 
We test \ppsam under two patient-adaptive segmentation tasks: NSCLC segmentation in the patient-adaptive radiation therapy and polyp segmentation in the endoscopy video. 
For NSCLC segmentation, medical image domain adaptation will be conducted on the \emph{internal} dataset, NSCLC-Radiomics. 
For \ppsam, experiments are then carried out on the \emph{external} dataset, 4D-Lung. 
We evaluate \ppsam on patients who underwent multiple visits during treatment. For each patient, we utilize the image-mask pair from the first visit as the patient-specific prior data. 
For polyp segmentation, domain adaptation will be conducted on \emph{internal} dataset, Kvasir-SEG. 
For \ppsam, experiments are then carried out on \emph{external} dataset, CVC-ClinicDB. For each video, we utilize the image-mask pair from the first stable frame as the patient-specific prior data.

\noindent\textbf{Implementation Details.} 
All experiments are conducted on A40 GPUs. 
For the NSCLC-Radiomics dataset, we extract 2-dimensional slices from the original computed tomography scans, resulting in a total of 7355 labeled images.
As for the Kvasir-SEG dataset, we utilize all 1000 labeled images. 
We process two datasets following existing works~\citep{8683802, dumitru2023using}. Each dataset was randomly split into three subsets: training, validation, and testing, with an $80\texttt{:}10\texttt{:}10$ percent ratio (patient-wise splitting for the NSCLC-Radiomics dataset to prevent data leak). 
The model is initialized with the SAM's pre-trained weights and fine-tuned on the training splitting using the loss function proposed by SAM. We optimize the model by AdamW optimizer~\citep{loshchilov2017decoupled}~($\beta_1 \texttt{=} 0.9, \beta_2 \texttt{=} 0.999$), with a weight decay of $0.05$. 
We further penalize the SAM's encoder with a drop path of $0.1$. 
We fine-tune the model for $36$ epochs on the NSCLC-Radiomics dataset and $100$ epochs on the Kvasir-SEG dataset with a batch size of $4$. 
The initial learning rate is $1e\texttt{-}4$, and the fine-tuning process is guided by cosine learning rate decay, with a linear learning rate warm-up over the first 10 percent epochs. 
More details are provided in Appendix~\ref{appendix: test implementation details}.

\noindent\textbf{Summary.} 
We test \ppsam on \emph{external} datasets with three different SAM backbones: 1. SAM pre-trained on the SA-1B dataset~\citep{kirillov2023segment}, denoted as \emph{Meta}. 2. SAM adapted on \emph{internal} datasets with LoRA~\citep{hu2021lora} and 3. full fine-tune, denoted as \emph{LoRA} and \emph{Full-Fine-Tune}, respectively. 
We compare \ppsam against various methods, including previous approaches such as the \emph{direct-transfer}; \emph{fine-tune} on the prior data~\citep{wang2019segmenting, elmahdy2020patient, wang2020predicting, chen2023patient}; the one-shot segmentation method, PANet~\citep{wang2019panet}; and concurrent methods that also utilize SAM, such as PerSAM~\citep{zhang2023personalize} and Matcher~\citep{liu2023matcher}. For PANet, we utilize its align method for one-shot segmentation. For Matcher, we adopt its setting of FSS-1000~\citep{li2020fss}. It is important to note that all baseline methods share the same backbone model as \ppsam does for fairness.

\subsection{Quantitative Results}
\label{subsec: quantitative results}

\begin{table}[tb]
    \caption{Results of NSCLC segmentation for patient-adaptive radiation therapy. We show the mean Dice score. \emph{base}$^\texttt{5.5M}$ indicates tuning \texttt{5.5M} parameters of the base SAM on the NSCLC-Radiomics dataset before testing on the 4D-Lung dataset. $\dagger$ indicates training-free method; $\ddagger$ indicates the method using SAM.}
    \begin{center}
    \setlength{\tabcolsep}{2pt}
    \begin{tabular}{c c c c c c}
        \toprule
        \multirow{2}*{Method} & \emph{Meta} & \multicolumn{2}{c}{\emph{LoRA}} & \multicolumn{2}{c}{\emph{Full-Fine-Tune}}\\
        \cmidrule(lr){2-2}\cmidrule(lr){3-4}\cmidrule(lr){5-6}
         \quad & \emph{huge}$^\texttt{0.0M}$ & \emph{base}$^\texttt{5.5M}$ & \emph{large}$^\texttt{5.9M}$ & \emph{base}$^\texttt{93.8M}$ & \emph{large}$^\texttt{312.5M}$\\
        \hline \vspace{-5pt}\\
        \emph{direct-transfer}$^{\dagger}$ & - & 56.10 & 57.83 & 58.18 & 61.11 \\
        \emph{fine-tune} & - & 52.11 & 32.55 & 55.27 & 53.85 \\
        PANet$^{\dagger}$~\citep{wang2019panet} & 4.28 & 5.24 & 7.79 & 40.03 & 44.70 \\
        \hline \vspace{-5pt} \\
        Matcher$^{\dagger\ddagger}$~\citep{liu2023matcher} & 13.28 & 50.81 & 50.88 & 59.52 & 57.67 \\
        PerSAM$^{\dagger\ddagger}$~\citep{zhang2023personalize} & 9.84 & 63.63 & 64.69 & 62.58 & 64.45 \\
        \ppsam$^{\dagger\ddagger}$~(Ours) & \bf 28.52 & \bf 64.38 & \bf 67.00 & \bf 66.68 & \bf 67.23 \\
        \bottomrule
    \end{tabular}
    \label{tab: result1}
    \end{center}
    \vspace{-10pt}
\end{table}

\begin{table}[tb]
    \caption{Results of polyp segmentation for endoscopy video. We show the mean Dice score for each method. \emph{base}$^\texttt{5.5M}$ indicates tuning \texttt{5.5M} parameters of the base SAM on the Kvasir-SEG dataset before testing on the CVC-ClinicDB dataset. $\dagger$ indicates training-free method; $\ddagger$ indicates the method using SAM.}
    \begin{center}
    \setlength{\tabcolsep}{2pt}
    \begin{tabular}{c c c c c c}
        \toprule
        \multirow{2}*{Method} & \emph{Meta} & \multicolumn{2}{c}{\emph{LoRA}} & \multicolumn{2}{c}{\emph{Full-Fine-Tune}}\\
        \cmidrule(lr){2-2}\cmidrule(lr){3-4}\cmidrule(lr){5-6}
         \quad & \emph{huge}$^\texttt{0.0M}$ & \emph{base}$^\texttt{5.5M}$ & \emph{large}$^\texttt{5.9M}$ & \emph{base}$^\texttt{93.8M}$ & \emph{large}$^\texttt{312.5M}$\\
        \hline \vspace{-5pt}\\
        \emph{direct-transfer}$^{\dagger}$ & - & 77.20 & 81.16 & 84.62 & 86.68 \\
        \emph{fine-tune} & - & 75.29 & 79.50 & 83.14 & 86.67 \\
        PANet$^{\dagger}$~\citep{wang2019panet} & 38.22 & 44.61 & 55.48 & 75.99 & 86.48 \\
        \hline \vspace{-5pt} \\
        Matcher$^{\dagger\ddagger}$~\citep{liu2023matcher} & 63.54 & 78.65 & 79.56 & 85.17 & 87.15 \\
        PerSAM$^{\dagger\ddagger}$~\citep{zhang2023personalize} & 45.82 & 79.02 & 81.63 & 85.74 & 87.88 \\
        \ppsam$^{\dagger\ddagger}$~(Ours) & \bf 66.45 & \bf 80.03 & \bf 82.60 & \bf 86.40 & \bf 88.76 \\
        \bottomrule
    \end{tabular}
    \label{tab: result2}
    \end{center}
    \vspace{-10pt}
\end{table}

\noindent\textbf{Patient-Adaptive Radiation Therapy.} 
As shown in \Tabref{tab: result1}, on the 4D-Lung dataset~\citep{hugo2016data}, \ppsam outperforms all other baselines across various backbones. 
Notably, when utilizing \emph{Meta}, \ppsam can outperform Matcher by $\texttt{+}15.24\%$ and PerSAM by $\texttt{+}18.68\%$ mean Dice score. This highlights \ppsam's superior adaptation to the out-of-domain medical applications.
After domain adaptation, \ppsam outperforms the \emph{direct-transfer} by $\texttt{+}8.01\%$, Matcher by $\texttt{+}11.60\%$, and PerSAM by $\texttt{+} 2.48\%$ mean Dice score, demonstrating that \ppsam is a more effective method to enhance generalization on the \emph{external} data.

\noindent\textbf{Discussion.}
\emph{fine-tune} is susceptible to overfitting with one-shot data, PANet fully depends on the encoder, and Matcher selects prompts based on patch-level features. These limitations prevent them from surpassing the \emph{direct-transfer}.
On the other hand, NSCLC segmentation remains a challenging task. We consider MedSAM~\citep{ma2024segment}, which has been pre-trained on a large-scale medical image dataset, as a strong \emph{baseline} method. In Table~\ref{tab: result3}, MedSAM achieves a 69\% mean dice score on the 4D-Lung dataset with a human-given box prompt at each visit, while \ppsam achieves comparable performance only with the ground truth provided at the first visit.

\noindent\textbf{Endoscopy Video.} 
As shown in \Tabref{tab: result2}, on the CVC-ClinicDB dataset~\citep{bernal2015wm}, \ppsam still achieves the best result across various backbones. 
When utilizing \emph{Meta}, \ppsam can surpass Matcher by $\texttt{+}2.91\%$ and PerSAM by $\texttt{+}20.63\%$ mean Dice score. 
After domain adaptation, \ppsam can outperform \emph{direct-transfer} by $\texttt{+}2.03\%$, Matcher by $\texttt{+}1.81\%$ and PerSAM by $\texttt{+}0.88\%$ mean Dice score. Demonstrates \ppsam's generality to various patient-adaptive segmentation tasks.

\noindent\textbf{Discussion.}
All methods demonstrate improved performance in datasets like CVC-ClinicDB, which exhibit a smaller domain gap~\citep{matsoukas2022makes} with SAM's pre-training dataset. In Table~\ref{tab: result3}, we compare our results with~\citet{sanderson2022fcn}, which is reported as the method achieving the best performance in~\citet{dumitru2023using} under the same evaluation objective: trained on Kvasir-SEG dataset and tested on the CVC-ClinicDB dataset. Our \emph{direct-transfer} has already surpassed this result, which can be attributed to the superior generality of SAM and our \ppsam can further improve the generalization. 

On the other hand, we observe that \ppsam's improvements over PerSAM become marginal after domain adaptation~(\emph{LoRA} and \emph{Full Fine-Tune} \emph{v.s.}~\emph{Meta}) on both datasets. This is because, as detailed in Appendix~\ref{appendix: sam adaptation details}, the ambiguity inherent in SAM, which is the primary limitation of PerSAM, is significantly reduced after fine-tuning on a dataset with a specific segmentation objective.
Nevertheless, our method shows that providing multiple curated prompts can achieve further improvement.

\begin{table}[t]
\noindent
\begin{minipage}[t]{0.42\textwidth}
    \caption{Comparison with existing baselines. $\star$ indicates using a human-given box prompt during the inference time.}
    \vspace{-1pt}
    \begin{center}
    \setlength{\tabcolsep}{2pt}
    \small
    \begin{tabular}{c c c}
        \toprule
        Method & 4D-Lung & CVC-ClinicDB \\
        \hline \vspace{-5pt} \\
        \emph{baseline} & 69.00$^\star$ & 83.14 \\
        \emph{direct-transfer} & 61.11 & 86.68 \\
        \ppsam & 67.23 & 88.76 \\
        \bottomrule
    \end{tabular}
    \vspace{-15pt}
    \label{tab: result3}
    \end{center}
\end{minipage}%
\hfill
\begin{minipage}[t]{0.54\textwidth}
    \caption{Results of one-shot semantic segmentation. We show the mean IoU score for each method. 
    % $\dagger$ indicates the training-free method. $\ddagger$ indicates the method using SAM.
    Note that all methods utilize SAM's encoder for fairness.
    }
    \begin{center}
    \setlength{\tabcolsep}{2pt}
    \small
    \begin{tabular}{c c c c c}
        \toprule
        Method 
        & COCO-$20^{i}$
        & FSS-1000
        & LVIS-$92^{i}$
        & PerSeg
        \\
        \hline \vspace{-5pt}\\
        Matcher
        & 25.1
        & 82.1 
        & 12.6
        & 90.2 
        \\
        PerSAM
        & 23.0
        & 71.2 
        & 11.5
        & 89.3 
        \\
        \ppsam~(Ours) 
        & \bf 26.0
        & \bf 82.4 
        & \bf 13.7
        & \bf 95.7 
        \\
        \bottomrule
    \end{tabular}
    \vspace{-15pt}
    \label{tab: result4}
    \end{center}
\end{minipage}
\end{table}

\begin{table}[tb]
\noindent
\begin{minipage}[t]{0.42\textwidth}
    \small
    \centering
    \setlength{\tabcolsep}{1pt}
    \caption{Comparison with tracking methods. $\ast$ indicates utilizing \emph{Full Fine-Tune}.}
    \vspace{5pt}
    \begin{tabular}{c c c}
    \toprule
    Method & 4D-Lung & CVC-ClinicDB \\
    \hline \vspace{-5pt}\\
    AOT & - & 62.34 \\
    \ppsam & - & 67.23 \\
    SAM~2 & - & 81.98 \\
    SAM~2 + \ppsam & - & \bf 84.43 \\
    \hline \vspace{-5pt}\\
    \emph{label-propagation}$^\ast$ & 57.00 & 82.92 \\
    \vspace{-7pt} \\
    \ppsam$^\ast$ & \bf 67.23 & \bf 88.76 \\
    \bottomrule
    \end{tabular}
    \label{tab: result5}
    \vspace{-15pt}
\end{minipage}%
\hfill
\begin{minipage}[t]{0.54\textwidth}
    \small
    \centering
    \setlength{\tabcolsep}{6pt}
    \caption{Ablation study for the number of parts $n$ and the retrieval. Default settings are marked in \colorbox{lightgray}{Gray}.}
    \begin{tabular}{c c c c c}
    \toprule
    \multirow{2}*{\# parts $(\mathit{n})$} & \multicolumn{2}{c}{CVC-ClinicDB} & \multicolumn{2}{c}{PerSeg}\\
    \cmidrule(lr){2-3}\cmidrule(lr){4-5}
    \quad & $\emph{w.o.}$ & $\emph{w.}$ retrieval & $\emph{w.o.}$ & $\emph{w.}$ retrieval \\
    \hline \vspace{-5pt}\\
    1 (PerSAM) & 45.8 & 45.8 & 89.3 & 89.3 \\
    \hline \vspace{-5pt}\\
    2 & 53.9 & 59.5 & 83.7 & 92.9 \\
    3 & 53.6 & 61.9 & 91.0 & 95.6 \\
    4 & 54.3 & 63.1 & 93.8 & 95.6 \\
    5 & 56.6 & \cellcolor{lightgray} 64.2 & 93.3 & \cellcolor{lightgray} 95.7 \\
    \bottomrule
    \end{tabular}
    \label{tab: ablation1}
    \vspace{-15pt}
\end{minipage}
\end{table}

\noindent\textbf{Comparison with Tracking Algorithms.} 
In Table~\ref{tab: result5}, we additionally compared \ppsam with tracking algorithms: the \emph{label-propagation}~\citep{jabri2020space}, AOT~\citep{yang2021associating}, and SAM~2~\citep{ravi2024sam}. On the 4D-Lung dataset, we only test algorithms with \emph{Full Fine-Tune} due to the large domain gap~\citep{matsoukas2022makes}. \ppsam outperforms the \emph{label-propagation}, as the discontinuity in sequential visits—where the interval between two CT scans can exceed a week—leads to significant changes in tumor position and features. On the CVC-ClinicDB dataset, dramatic content shifts within the narrow field of view can also lead to discontinuity. Despite this, SAM~2 achieves competitive results even without additional domain adaptation. However, as we have stated, \ppsam can be integrated into any promptable segmentation model. Indeed, we observe further improvements when applying \ppsam to SAM~2.

\noindent\textbf{Existing One-shot Segmentation Benchmarks.}
To further demonstrate \ppsam can also be generalized to natural image domain, we evaluate its performance on existing one-shot semantic segmentation benchmarks: COCO-$20^{i}$~\citep{nguyen2019feature}, FSS-1000~\citep{li2020fss}, LVIS-$92^{i}$~\citep{liu2023matcher}, and a personalized segmentation benchmark, PerSeg~\citep{zhang2023personalize}. We follow previous works~\citep{zhang2023personalize, liu2023matcher} for data pre-processing and evaluation.
In \Tabref{tab: result4}, when utilizing SAM's encoder, \ppsam outperforms concurrent works, Matcher and PerSAM, on all existing benchmarks. In addition, \ppsam can achieve a new state-of-the-art result, $95.7\%$ mean IoU score, on the personalized segmentation benchmark PerSeg~\citep{zhang2023personalize}.

\begin{figure*}[t]
    \vspace{-15pt}
    \noindent
    \begin{minipage}[t]{.48\textwidth}
        \centering
        \includegraphics[width=\linewidth]{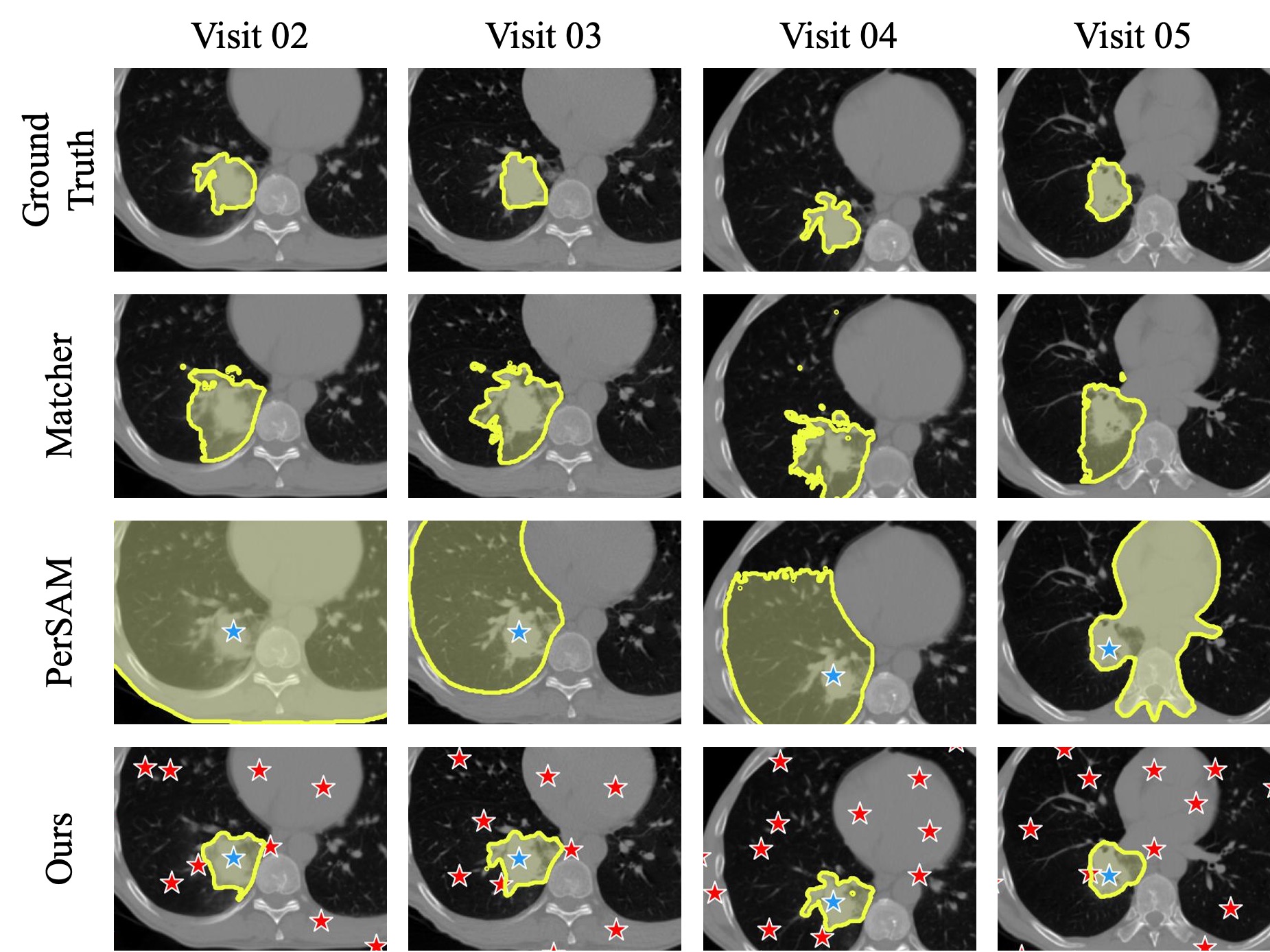} 
        \vspace{-15pt}
        \caption{Qualitative results of NSCLC segmentation on the 4D-Lung dataset, with \emph{Meta}.} 
        \label{fig: result1-1}
    \end{minipage}%
    \vspace{2pt}
    \hfill
    \begin{minipage}[t]{.48\textwidth} 
        \centering
        \includegraphics[width=\linewidth]{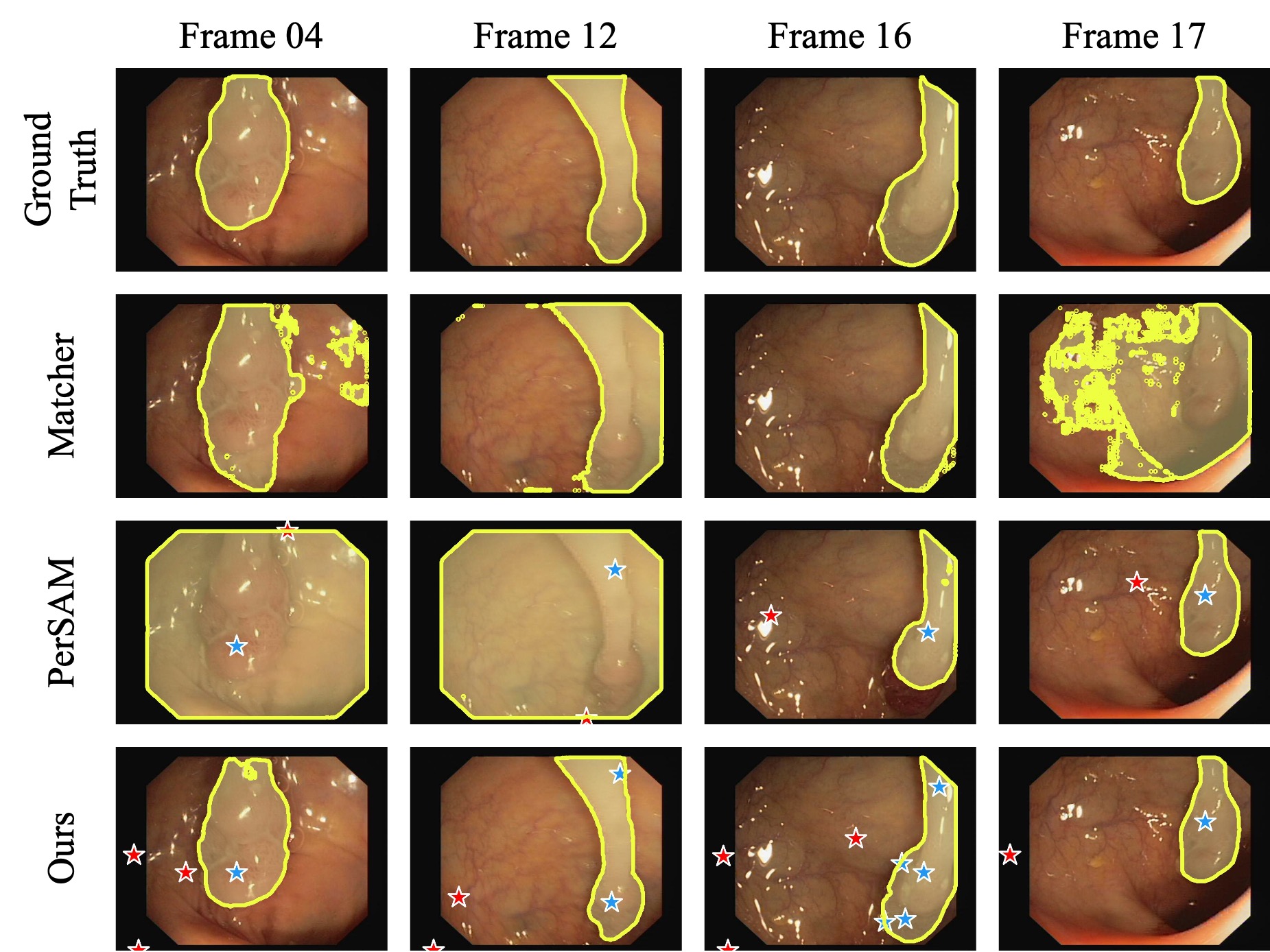}
        \vspace{-15pt}
        \caption{Qualitative results of polyp segmentation on the CVC-ClinicDB dataset, with \emph{Meta}.}
        \label{fig: result1-2}
    \end{minipage}%
    \vspace{2pt}
    \\
    \noindent
    \begin{minipage}[t]{.48\textwidth}
        \centering
        \includegraphics[width=\linewidth]{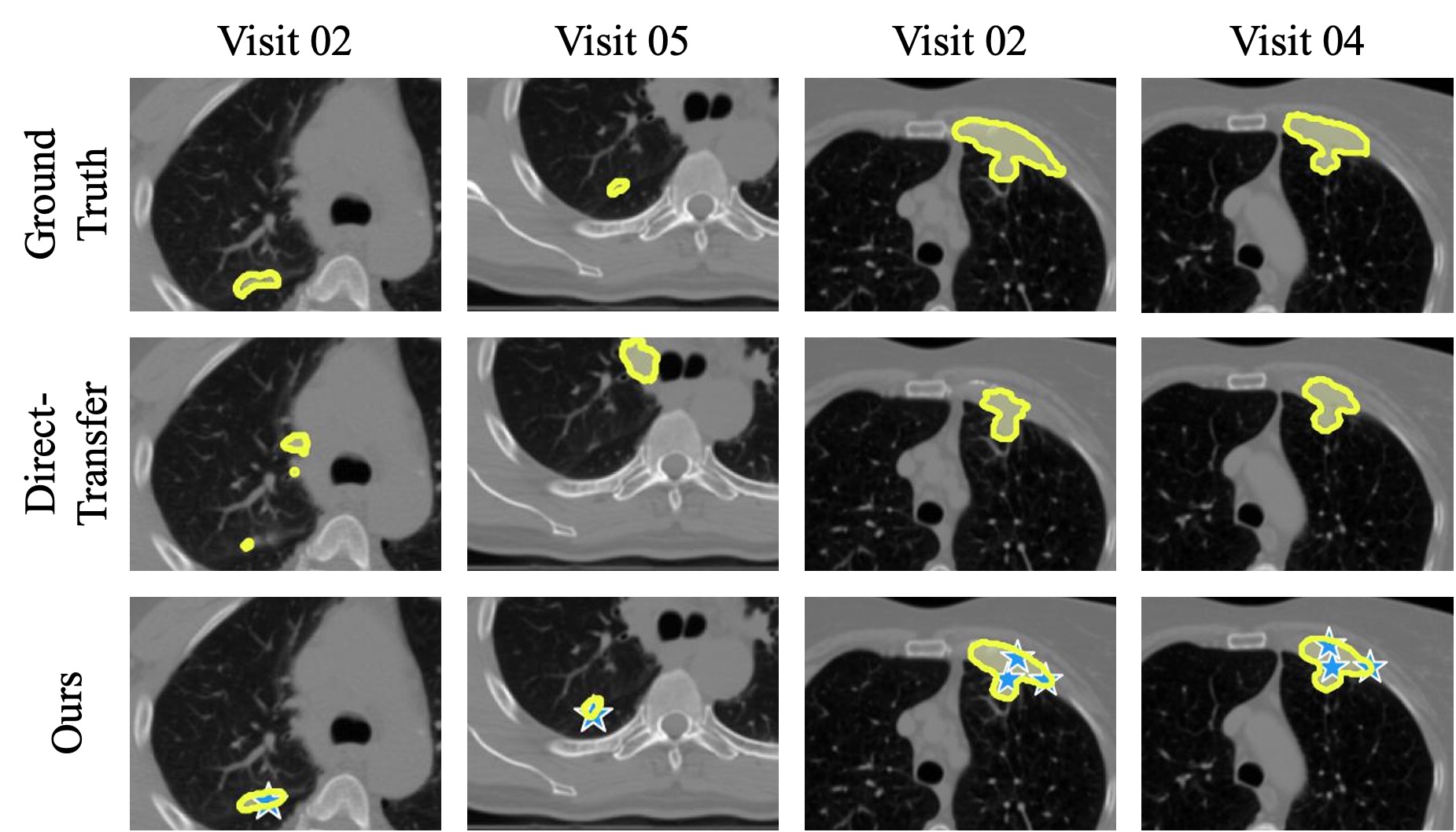} 
        \vspace{-15pt}
        \caption{Qualitative results of NSCLC segmentation from two patients on the 4D-Lung dataset, with \emph{Full-Fine-Tune}.} 
        \label{fig: result2-1}
    \end{minipage}%
    \vspace{2pt}
    \hfill
    \begin{minipage}[t]{.48\textwidth} 
        \centering
        \includegraphics[width=\linewidth]{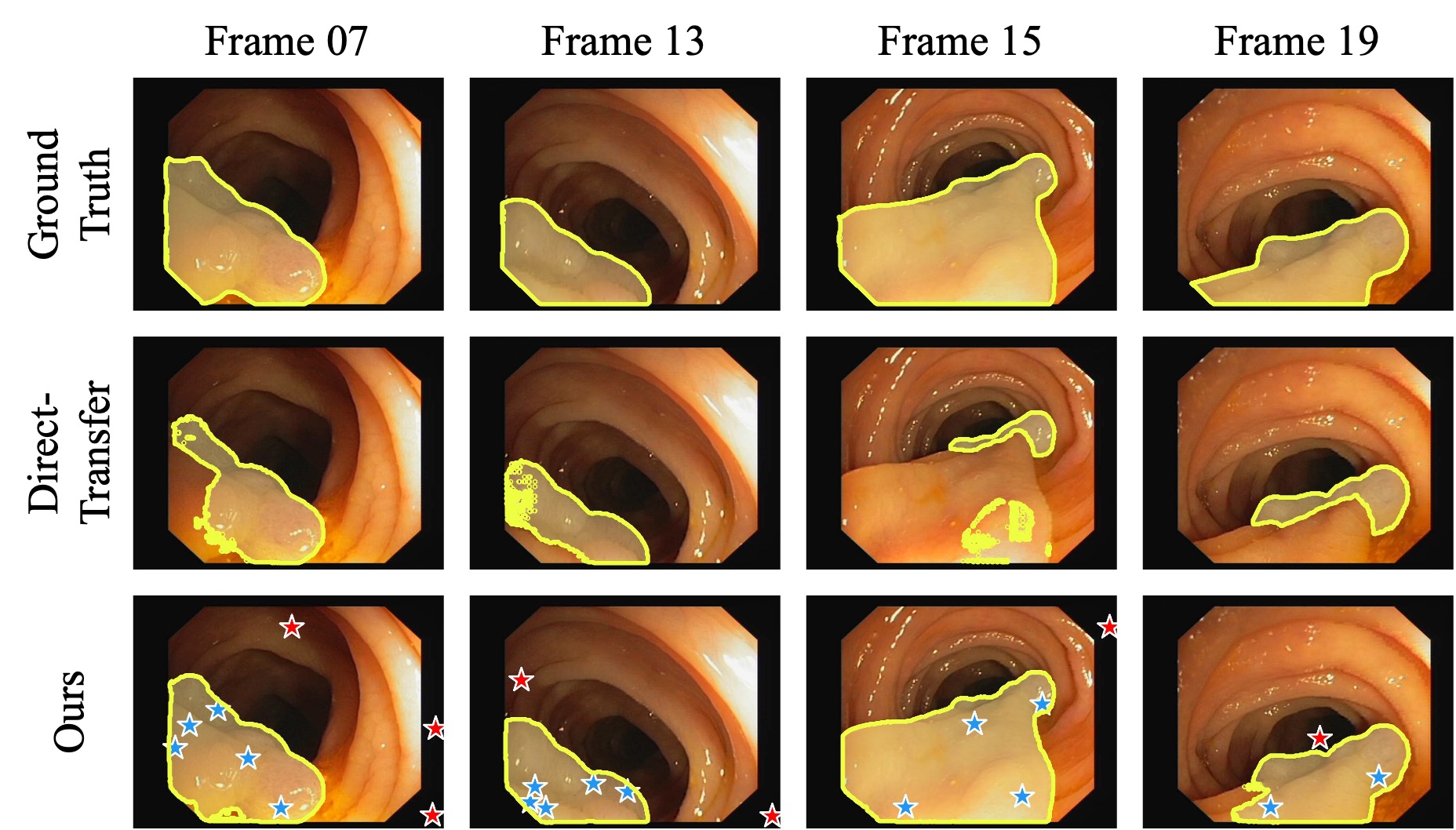}
        \vspace{-15pt}
        \caption{Qualitative results of polyp segmentation from one video on the CVC-ClinicDB dataset, with \emph{Full-Fine-Tune}.}
        \label{fig: result2-2}
    \end{minipage}%
    \vspace{2pt}
    \\
    \noindent
    \begin{minipage}[t]{.48\textwidth}
        \centering
        \includegraphics[width=\linewidth]{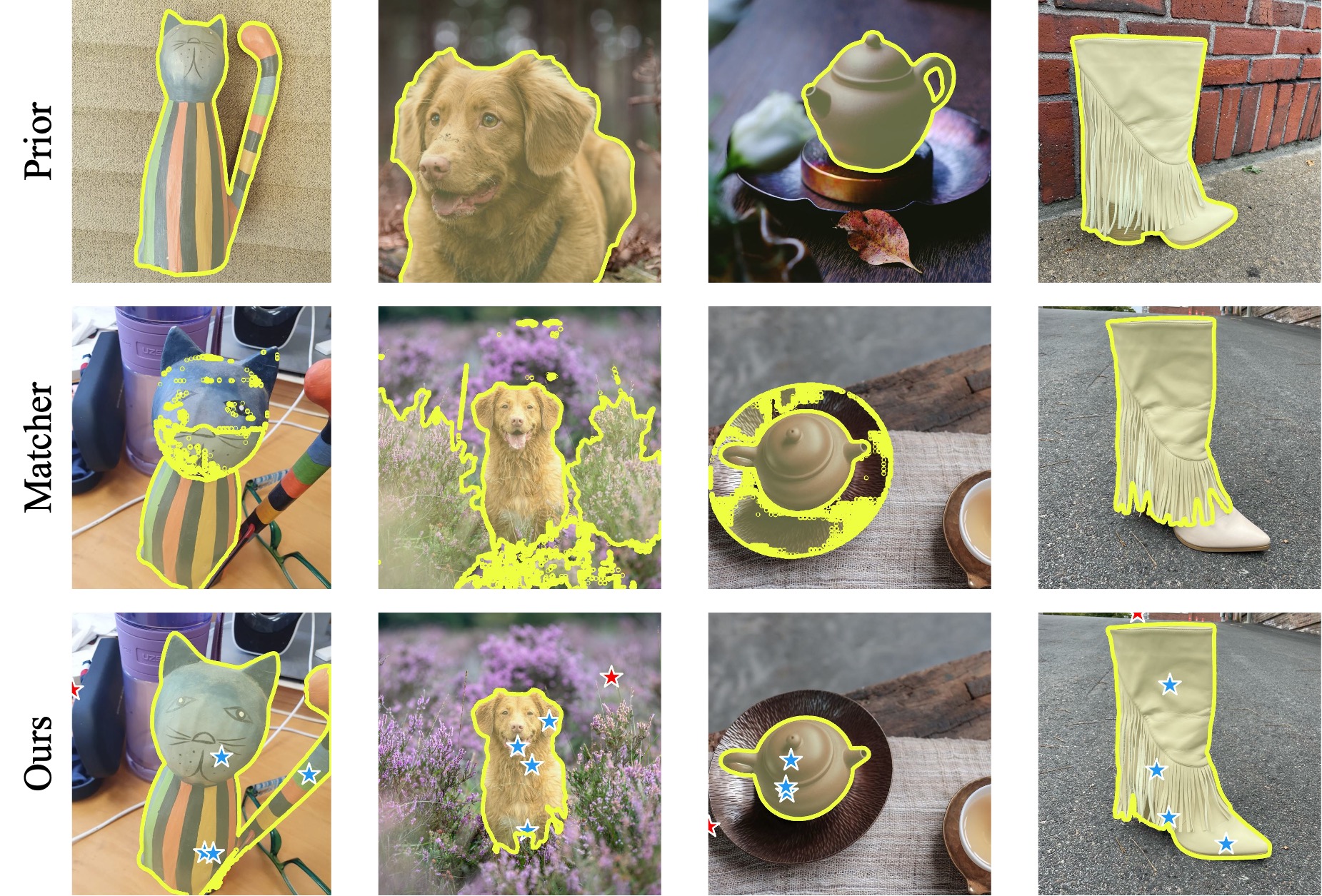} 
        \vspace{-15pt}
        \caption{Qualitative results of personalized segmentation on the PerSeg dataset, compared with Matcher.} 
        \label{fig: result3-1}
    \end{minipage}%
    \hfill
    \begin{minipage}[t]{.48\textwidth} 
        \centering
        \includegraphics[width=\linewidth]{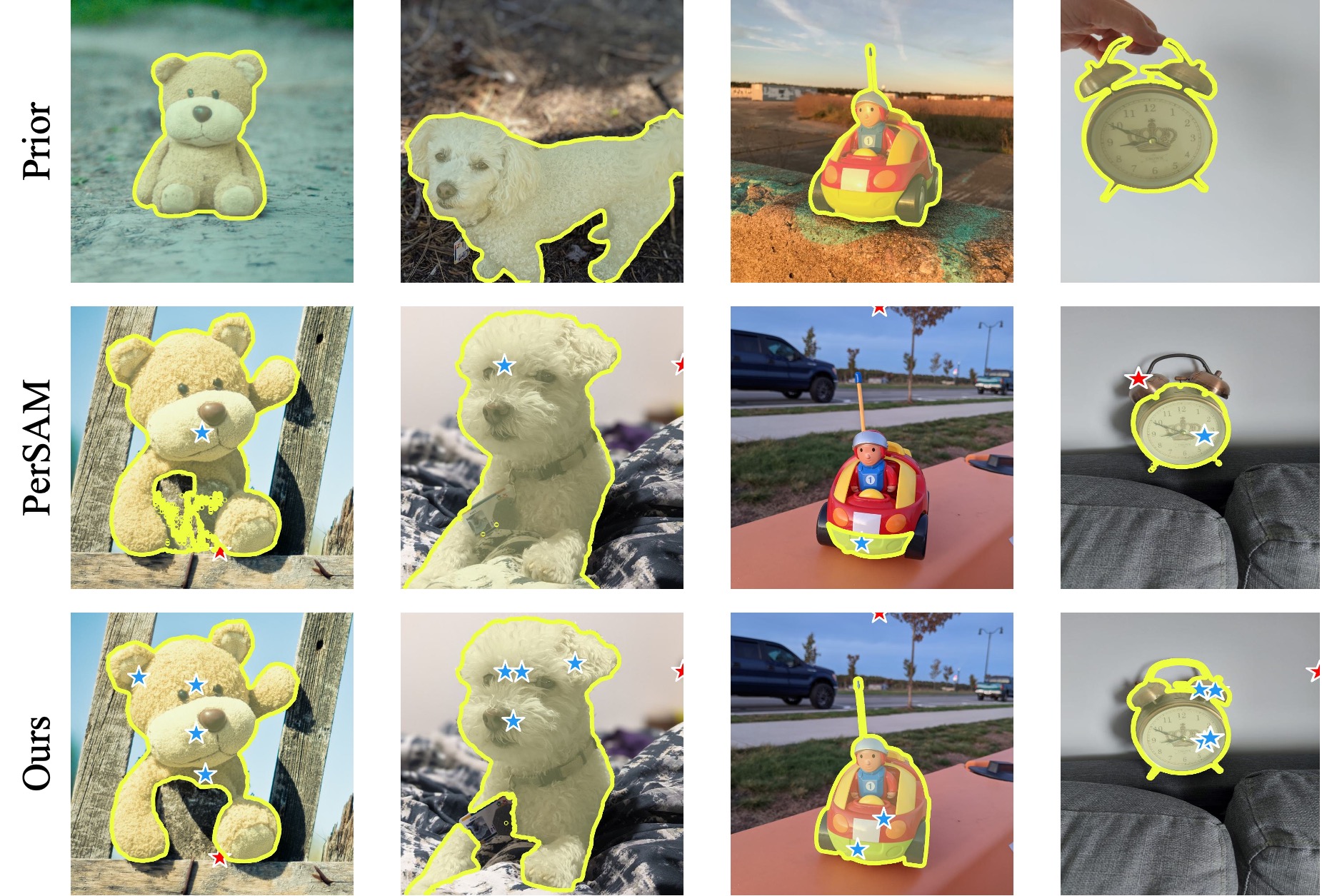}
        \vspace{-15pt}
        \caption{Qualitative results of personalized segmentation on the PerSeg dataset, compared with PerSAM.}
        \label{fig: result3-2}
    \end{minipage}%
    \vspace{-15pt}
\end{figure*}

\subsection{Ablation Study}
\label{subsec: ablation study}

\begin{table}[t]
\noindent
\begin{minipage}[t]{0.42\textwidth}
    \small
    \centering
    \setlength{\tabcolsep}{4pt}
    \caption{Ablation study for the distribution distance measurement. Default settings are marked in \colorbox{lightgray}{Gray}.}
    \begin{tabular}{c c c}
    \toprule
    Algorithm & CVC-ClinicDB & PerSeg \\
    \hline \vspace{-4pt}\\
    \emph{w.o.} & 54.3 & 93.8 \\
    \hline \vspace{-4pt}\\
    \textit{Hungarian} & 61.1 & 95.6 \\
    \vspace{1pt}
    \textit{Jensen–Shannon} & 58.1 & 94.0 \\
    \rowcolor{lightgray}
    \vspace{1pt}
    \textit{Wasserstein} & 63.1 & 95.6 \\
    \bottomrule
    \end{tabular}
    \label{tab: ablation2}
    \vspace{-10pt}
\end{minipage}%
\hfill
\begin{minipage}[t]{0.54\textwidth}
    \small
    \centering
    \setlength{\tabcolsep}{8pt}
    \caption{Ablation study for model sizes. ${\uparrow}$ indicates the improvement when compared with the same size PerSAM. Default settings are marked in \colorbox{lightgray}{Gray}.}
    \begin{tabular}{ccc}
    \toprule
    Model & CVC-ClinicDB & PerSeg \\
    \hline \vspace{-5pt}\\
    PerSAM$^{huge}$ & 45.8 & 89.3 \\
    \hline \vspace{-5pt}\\
    \ppsam$^{base}$ & 55.1 & 90.0 ${_{\texttt{26.0}\uparrow}}$ \\
    \ppsam$^{large}$ & 63.8 & 95.6 ${_{\texttt{9.0}\uparrow}}$ \\
    \rowcolor{lightgray}
    \ppsam$^{huge}$ & 63.1 & 95.6 ${_{\texttt{6.3}\uparrow}}$ \\
    \bottomrule
    \end{tabular}
    \label{tab: ablation3}
    \vspace{-10pt}
\end{minipage}
\end{table}

Ablation studies are conducted on the PerSeg dataset~\citep{zhang2023personalize} and CVC-ClinicDB dataset~\citep{bernal2015wm} using \emph{Meta}. We explore the effects of the number of parts in the part-aware prompt mechanism; the retrieval approach; distribution distance measurements in the retrieval approach; and the model size, which can be considered a proxy for representation capacity.

\noindent\textbf{Number of Parts \textit{n}.}
To validate the efficacy of the part-aware prompt mechanism, we establish a method without the retrieval approach. As shown in \Tabref{tab: ablation1}~(\emph{w.o.} retrieval), for both datasets, even solely relying on the part-aware prompt mechanism, increasing the number of parts $n$ enhances segmentation performance.
When setting $n\texttt{=}5$, our part-aware prompt mechanism enhances performance by $\texttt{+}10.7\%$ mean Dice score on CVC-ClinicDB, $\texttt{+}4.0\%$ mean IoU score on PerSeg. These substantial improvements underscore the effectiveness of our part-aware prompt mechanism.

\noindent\textbf{Retrieval Approach.}
The effectiveness of our retrieval approach is also shown in \Tabref{tab: ablation1}~(\emph{w.} retrieval). 
When setting $n\texttt{=}5$, the retrieval approach enhances performance by $\texttt{+}7.6\%$ mean Dice score on the CVC-ClinicDB dataset, $\texttt{+}2.4\%$ mean IoU score on the PerSeg dataset. 
These substantial improvements show that our retrieval approach can retrieve an appropriate number of parts for different cases.
Moreover, these suggest that we can initially define a wide range of part counts for retrieval, rather than tuning it meticulously as a hyperparameter.

\noindent\textbf{Distribution Distance Measurements.}
The cornerstone of our retrieval approach lies in distribution distance measurements. To evaluate the efficacy of various algorithms, in \Tabref{tab: ablation2}, we juxtapose two distribution-related algorithms, namely \emph{Wasserstein} distance~\citep{ruschendorf1985wasserstein} and \emph{Jensen–Shannon} divergence~\citep{menendez1997jensen}, alongside a bipartite matching algorithm, \emph{Hungarian} algorithm.
Given foreground features from the reference image and the target image, we compute: 1. \emph{Wasserstein} distance following the principles of WGAN~\citep{arjovsky2017wasserstein}; 2. \emph{Jensen-Shannon} divergence based on the first two principal components of each feature;
3. \emph{Hungarian} algorithm after clustering these two sets of features into an equal number of parts.
All algorithms exhibit improvements in segmentation performance compared to the \emph{w.o.} retrieval baseline, while the \emph{Wasserstein} distance is better in our context. Note that, the efficacy of the \emph{Jensen-Shannon} divergence further corroborates our assumption that foreground features from the reference image and a correct target result should align in the same distribution, albeit it faces challenges when handling the high-dimensional data.

\noindent\textbf{Model Size.}
In \Tabref{tab: ablation3}, we investigate the performance of different model sizes for our \ppsam, \ie, \emph{base}, \emph{large}, and \emph{huge}, which can alternatively be viewed as the representation capacity of different backbones. 
For the CVC-ClinicDB dataset, a larger model size does not necessarily lead to better results. This result aligns with current conclusions~\citep{mazurowski2023segment, huang2024segment}: In medical image analysis, the \emph{huge} SAM may occasionally be outperformed by the \emph{large} SAM. 
On the other hand, for the PerSeg dataset, even utilizing the \emph{base} SAM, \ppsam achieves higher accuracy compared to PerSAM with the \emph{huge} SAM. 
These findings further underscore the robustness of \ppsam, particularly in scenarios where the model exhibits weaker representation, a circumstance more prevalent in medical image analysis.

\subsection{Qualitative Results}
\label{subsec: qualitative results}

Figure~\ref{fig: result1-1} and~\ref{fig: result1-2} showcase the advantage of \ppsam for out-of-domain applications.
As shown in \Figref{fig: result1-1}, by presenting sufficient negative-point prompts, we enforce the model's focus on the semantic target. 
Results in \Figref{fig: result1-2} further summarize the benefits of our method: unambiguous segmentation and robust prompts selection. 
Our \ppsam can also improve the model's generalization after domain adaptation. By providing precise foreground information, \ppsam enhances segmentation performance when the object is too small~(\eg, the first two columns in Figure~\ref{fig: result2-1}) and when the segmentation is incomplete~(\eg, the last two columns in Figure~\ref{fig: result2-2}).
\Figref{fig: result3-1} and~\ref{fig: result3-2} showcase the qualitative results on the PerSeg dataset, compared with Matcher and PerSAM respectively. The remarkable results demonstrate that \ppsam can generalize well to different domain applications. 

\section{Conclusion}
\label{sec: conclusion}

We propose a data-efficient segmentation method, \ppsam, to solve the patient-adaptive segmentation problem. With a novel part-aware prompt mechanism and a distribution-guided retrieval approach, \ppsam can effectively integrate the patient-specific prior information into the current segmentation task. 
Beyond patient-adaptive segmentation, \ppsam demonstrates promising versatility in enhancing the backbone's generalization across various levels: 
1. At the domain level, \ppsam performs effectively in both medical and natural image domains.
2. At the task level, \ppsam enhances performance across different patient-adaptive segmentation tasks. 
3. At the model level, \ppsam can be integrated into various promptable segmentation models, such as SAM, SAM~2, and custom fine-tuned SAM. 
In this work, to meet clinical requirements, we choose to adapt SAM to the medical imaging domain with public datasets. We opted not to adopt SAM~2, as it requires video data for fine-tuning, which is more costly. Additionally, treating certain patient-adaptive segmentation tasks as video tracking is inappropriate. In contrast, approaching patient-adaptive segmentation as an in-context segmentation problem offers a more flexible solution for various patient-adaptive segmentation tasks.
Additional discussions can be found in appendix.
We hope our work brings attention to the patient-adaptive segmentation problem within the research community.

\section*{Acknowledgments}
The authors acknowledge support from University of Michigan MIDAS (Michigan Institute for Data Science) PODS Grant and University of Michigan MICDE (Michigan Institute for Computational Discovery and Engineering) Catalyst Grant, and the computing resource support from NSF ACCESS Program.

\clearpage
\bibliography{main}

\begin{thebibliography}{70}
\providecommand{\natexlab}[1]{#1}
\providecommand{\url}[1]{\texttt{#1}}
\expandafter\ifx\csname urlstyle\endcsname\relax
  \providecommand{\doi}[1]{doi: #1}\else
  \providecommand{\doi}{doi: \begingroup \urlstyle{rm}\Url}\fi

\bibitem[Aerts et~al.(2015)Aerts, Velazquez, Leijenaar, Parmar, Grossmann, Cavalho, Bussink, Monshouwer, Haibe-Kains, Rietveld, et~al.]{aerts2015data}
HJWL Aerts, E~Rios Velazquez, RT~Leijenaar, Chintan Parmar, Patrick Grossmann, S~Cavalho, Johan Bussink, Ren{\'e} Monshouwer, Benjamin Haibe-Kains, Derek Rietveld, et~al.
\newblock Data from nsclc-radiomics.
\newblock \emph{The cancer imaging archive}, 2015.

\bibitem[Antonelli et~al.(2022)Antonelli, Reinke, Bakas, Farahani, Kopp-Schneider, Landman, Litjens, Menze, Ronneberger, Summers, et~al.]{antonelli2022medical}
Michela Antonelli, Annika Reinke, Spyridon Bakas, Keyvan Farahani, Annette Kopp-Schneider, Bennett~A Landman, Geert Litjens, Bjoern Menze, Olaf Ronneberger, Ronald~M Summers, et~al.
\newblock The medical segmentation decathlon.
\newblock \emph{Nature communications}, 13\penalty0 (1):\penalty0 4128, 2022.

\bibitem[Arjovsky et~al.(2017)Arjovsky, Chintala, and Bottou]{arjovsky2017wasserstein}
Martin Arjovsky, Soumith Chintala, and L{\'e}on Bottou.
\newblock Wasserstein generative adversarial networks.
\newblock In \emph{International conference on machine learning}, pp.\  214--223. PMLR, 2017.

\bibitem[Arthur et~al.(2007)Arthur, Vassilvitskii, et~al.]{arthur2007k}
David Arthur, Sergei Vassilvitskii, et~al.
\newblock k-means++: The advantages of careful seeding.
\newblock In \emph{Soda}, volume~7, pp.\  1027--1035, 2007.

\bibitem[Bernal et~al.(2015)Bernal, S{\'a}nchez, Fern{\'a}ndez-Esparrach, Gil, Rodr{\'\i}guez, and Vilari{\~n}o]{bernal2015wm}
Jorge Bernal, F~Javier S{\'a}nchez, Gloria Fern{\'a}ndez-Esparrach, Debora Gil, Cristina Rodr{\'\i}guez, and Fernando Vilari{\~n}o.
\newblock Wm-dova maps for accurate polyp highlighting in colonoscopy: Validation vs. saliency maps from physicians.
\newblock \emph{Computerized medical imaging and graphics}, 43:\penalty0 99--111, 2015.

\bibitem[Brown et~al.(2020)Brown, Mann, Ryder, Subbiah, Kaplan, Dhariwal, Neelakantan, Shyam, Sastry, Askell, et~al.]{brown2020language}
Tom Brown, Benjamin Mann, Nick Ryder, Melanie Subbiah, Jared~D Kaplan, Prafulla Dhariwal, Arvind Neelakantan, Pranav Shyam, Girish Sastry, Amanda Askell, et~al.
\newblock Language models are few-shot learners.
\newblock \emph{Advances in neural information processing systems}, 33:\penalty0 1877--1901, 2020.

\bibitem[Butoi et~al.(2023)Butoi, Ortiz, Ma, Sabuncu, Guttag, and Dalca]{butoi2023universeg}
Victor~Ion Butoi, Jose Javier~Gonzalez Ortiz, Tianyu Ma, Mert~R Sabuncu, John Guttag, and Adrian~V Dalca.
\newblock Universeg: Universal medical image segmentation.
\newblock In \emph{Proceedings of the IEEE/CVF International Conference on Computer Vision}, pp.\  21438--21451, 2023.

\bibitem[Carion et~al.(2020)Carion, Massa, Synnaeve, Usunier, Kirillov, and Zagoruyko]{carion2020end}
Nicolas Carion, Francisco Massa, Gabriel Synnaeve, Nicolas Usunier, Alexander Kirillov, and Sergey Zagoruyko.
\newblock End-to-end object detection with transformers.
\newblock In \emph{European conference on computer vision}, pp.\  213--229. Springer, 2020.

\bibitem[Chen et~al.(2023)Chen, Gensheimer, Bagshaw, Butler, Yu, Zhou, Shen, Kovalchuk, Surucu, Chang, et~al.]{chen2023patient}
Yizheng Chen, Michael~F Gensheimer, Hilary~P Bagshaw, Santino Butler, Lequan Yu, Yuyin Zhou, Liyue Shen, Nataliya Kovalchuk, Murat Surucu, Daniel~T Chang, et~al.
\newblock Patient-specific auto-segmentation on daily kvct images for adaptive radiotherapy.
\newblock \emph{International Journal of Radiation Oncology* Biology* Physics}, 2023.

\bibitem[Cheng et~al.(2021)Cheng, Schwing, and Kirillov]{cheng2021per}
Bowen Cheng, Alex Schwing, and Alexander Kirillov.
\newblock Per-pixel classification is not all you need for semantic segmentation.
\newblock \emph{Advances in Neural Information Processing Systems}, 34:\penalty0 17864--17875, 2021.

\bibitem[Dosovitskiy et~al.(2020)Dosovitskiy, Beyer, Kolesnikov, Weissenborn, Zhai, Unterthiner, Dehghani, Minderer, Heigold, Gelly, et~al.]{dosovitskiy2020image}
Alexey Dosovitskiy, Lucas Beyer, Alexander Kolesnikov, Dirk Weissenborn, Xiaohua Zhai, Thomas Unterthiner, Mostafa Dehghani, Matthias Minderer, Georg Heigold, Sylvain Gelly, et~al.
\newblock An image is worth 16x16 words: Transformers for image recognition at scale.
\newblock \emph{arXiv preprint arXiv:2010.11929}, 2020.

\bibitem[Dumitru et~al.(2023)Dumitru, Peteleaza, and Craciun]{dumitru2023using}
Razvan-Gabriel Dumitru, Darius Peteleaza, and Catalin Craciun.
\newblock Using duck-net for polyp image segmentation.
\newblock \emph{Scientific Reports}, 13\penalty0 (1):\penalty0 9803, 2023.

\bibitem[Elmahdy et~al.(2020)Elmahdy, Ahuja, van~der Heide, and Staring]{elmahdy2020patient}
Mohamed~S Elmahdy, Tanuj Ahuja, Uulke~A van~der Heide, and Marius Staring.
\newblock Patient-specific finetuning of deep learning models for adaptive radiotherapy in prostate ct.
\newblock In \emph{2020 IEEE 17th International Symposium on Biomedical Imaging (ISBI)}, pp.\  577--580. IEEE, 2020.

\bibitem[Everingham et~al.(2010)Everingham, Van~Gool, Williams, Winn, and Zisserman]{everingham2010pascal}
Mark Everingham, Luc Van~Gool, Christopher~KI Williams, John Winn, and Andrew Zisserman.
\newblock The pascal visual object classes (voc) challenge.
\newblock \emph{International journal of computer vision}, 88:\penalty0 303--338, 2010.

\bibitem[Garc{\'\i}a-Figueiras et~al.(2019)Garc{\'\i}a-Figueiras, Baleato-Gonz{\'a}lez, Padhani, Luna-Alcal{\'a}, Vallejo-Casas, Sala, Vilanova, Koh, Herranz-Carnero, and Vargas]{garcia2019clinical}
Roberto Garc{\'\i}a-Figueiras, Sandra Baleato-Gonz{\'a}lez, Anwar~R Padhani, Antonio Luna-Alcal{\'a}, Juan~Antonio Vallejo-Casas, Evis Sala, Joan~C Vilanova, Dow-Mu Koh, Michel Herranz-Carnero, and Herbert~Alberto Vargas.
\newblock How clinical imaging can assess cancer biology.
\newblock \emph{Insights into imaging}, 10:\penalty0 1--35, 2019.

\bibitem[He et~al.(2017)He, Gkioxari, Doll{\'a}r, and Girshick]{he2017mask}
Kaiming He, Georgia Gkioxari, Piotr Doll{\'a}r, and Ross Girshick.
\newblock Mask r-cnn.
\newblock In \emph{Proceedings of the IEEE international conference on computer vision}, pp.\  2961--2969, 2017.

\bibitem[Hodson(2016)]{hodson2016precision}
Richard Hodson.
\newblock Precision medicine.
\newblock \emph{Nature}, 537\penalty0 (7619):\penalty0 S49--S49, 2016.

\bibitem[Hossain et~al.(2019)Hossain, Najeeb, Shahriyar, Abdullah, and Ariful~Haque]{8683802}
Shahruk Hossain, Suhail Najeeb, Asif Shahriyar, Zaowad~R. Abdullah, and M.~Ariful~Haque.
\newblock A pipeline for lung tumor detection and segmentation from ct scans using dilated convolutional neural networks.
\newblock In \emph{ICASSP 2019 - 2019 IEEE International Conference on Acoustics, Speech and Signal Processing (ICASSP)}, pp.\  1348--1352, 2019.
\newblock \doi{10.1109/ICASSP.2019.8683802}.

\bibitem[Hu et~al.(2021)Hu, Shen, Wallis, Allen-Zhu, Li, Wang, Wang, and Chen]{hu2021lora}
Edward~J Hu, Yelong Shen, Phillip Wallis, Zeyuan Allen-Zhu, Yuanzhi Li, Shean Wang, Lu~Wang, and Weizhu Chen.
\newblock Lora: Low-rank adaptation of large language models.
\newblock \emph{arXiv preprint arXiv:2106.09685}, 2021.

\bibitem[Huang et~al.(2024)Huang, Yang, Liu, Zhou, Chang, Zhou, Chen, Yu, Chen, Chen, et~al.]{huang2024segment}
Yuhao Huang, Xin Yang, Lian Liu, Han Zhou, Ao~Chang, Xinrui Zhou, Rusi Chen, Junxuan Yu, Jiongquan Chen, Chaoyu Chen, et~al.
\newblock Segment anything model for medical images?
\newblock \emph{Medical Image Analysis}, 92:\penalty0 103061, 2024.

\bibitem[Hugo et~al.(2016)Hugo, Weiss, Sleeman, Balik, Keall, Lu, and Williamson]{hugo2016data}
Geoffrey~D Hugo, Elisabeth Weiss, William~C Sleeman, Salim Balik, Paul~J Keall, Jun Lu, and Jeffrey~F Williamson.
\newblock Data from 4d lung imaging of nsclc patients.
\newblock \emph{The Cancer Imaging Archive}, 10:\penalty0 K9, 2016.

\bibitem[Isensee et~al.(2021)Isensee, Jaeger, Kohl, Petersen, and Maier-Hein]{isensee2021nnu}
Fabian Isensee, Paul~F Jaeger, Simon~AA Kohl, Jens Petersen, and Klaus~H Maier-Hein.
\newblock nnu-net: a self-configuring method for deep learning-based biomedical image segmentation.
\newblock \emph{Nature methods}, 18\penalty0 (2):\penalty0 203--211, 2021.

\bibitem[Jabri et~al.(2020)Jabri, Owens, and Efros]{jabri2020space}
Allan Jabri, Andrew Owens, and Alexei Efros.
\newblock Space-time correspondence as a contrastive random walk.
\newblock \emph{Advances in neural information processing systems}, 33:\penalty0 19545--19560, 2020.

\bibitem[Jha et~al.(2020)Jha, Smedsrud, Riegler, Halvorsen, de~Lange, Johansen, and Johansen]{jha2020kvasir}
Debesh Jha, Pia~H Smedsrud, Michael~A Riegler, P{\aa}l Halvorsen, Thomas de~Lange, Dag Johansen, and H{\aa}vard~D Johansen.
\newblock Kvasir-seg: A segmented polyp dataset.
\newblock In \emph{MultiMedia Modeling: 26th International Conference, MMM 2020, Daejeon, South Korea, January 5--8, 2020, Proceedings, Part II 26}, pp.\  451--462. Springer, 2020.

\bibitem[Ji et~al.(2022)Ji, Bai, Ge, Yang, Zhu, Zhang, Li, Zhanng, Ma, Wan, et~al.]{ji2022amos}
Yuanfeng Ji, Haotian Bai, Chongjian Ge, Jie Yang, Ye~Zhu, Ruimao Zhang, Zhen Li, Lingyan Zhanng, Wanling Ma, Xiang Wan, et~al.
\newblock Amos: A large-scale abdominal multi-organ benchmark for versatile medical image segmentation.
\newblock \emph{Advances in neural information processing systems}, 35:\penalty0 36722--36732, 2022.

\bibitem[Kirillov et~al.(2023)Kirillov, Mintun, Ravi, Mao, Rolland, Gustafson, Xiao, Whitehead, Berg, Lo, et~al.]{kirillov2023segment}
Alexander Kirillov, Eric Mintun, Nikhila Ravi, Hanzi Mao, Chloe Rolland, Laura Gustafson, Tete Xiao, Spencer Whitehead, Alexander~C Berg, Wan-Yen Lo, et~al.
\newblock Segment anything.
\newblock \emph{arXiv preprint arXiv:2304.02643}, 2023.

\bibitem[Leng et~al.(2024)Leng, Zhang, Han, and Xie]{leng2024self}
Tianang Leng, Yiming Zhang, Kun Han, and Xiaohui Xie.
\newblock Self-sampling meta sam: Enhancing few-shot medical image segmentation with meta-learning.
\newblock In \emph{Proceedings of the IEEE/CVF Winter Conference on Applications of Computer Vision}, pp.\  7925--7935, 2024.

\bibitem[Li et~al.(2020)Li, Wei, Chen, Tai, and Tang]{li2020fss}
Xiang Li, Tianhan Wei, Yau~Pun Chen, Yu-Wing Tai, and Chi-Keung Tang.
\newblock Fss-1000: A 1000-class dataset for few-shot segmentation.
\newblock In \emph{Proceedings of the IEEE/CVF conference on computer vision and pattern recognition}, pp.\  2869--2878, 2020.

\bibitem[Li et~al.(2022{\natexlab{a}})Li, Wang, Li, and Lu]{li2022hybrid}
Xiang Li, Jinglu Wang, Xiao Li, and Yan Lu.
\newblock Hybrid instance-aware temporal fusion for online video instance segmentation.
\newblock In \emph{Proceedings of the AAAI Conference on Artificial Intelligence}, volume~36, pp.\  1429--1437, 2022{\natexlab{a}}.

\bibitem[Li et~al.(2023{\natexlab{a}})Li, Lin, Chen, Liu, Wang, and Raj]{li2023paintseg}
Xiang Li, Chung-Ching Lin, Yinpeng Chen, Zicheng Liu, Jinglu Wang, and Bhiksha Raj.
\newblock Paintseg: Training-free segmentation via painting.
\newblock \emph{arXiv preprint arXiv:2305.19406}, 2023{\natexlab{a}}.

\bibitem[Li et~al.(2023{\natexlab{b}})Li, Wang, Xu, Li, Raj, and Lu]{li2023robust}
Xiang Li, Jinglu Wang, Xiaohao Xu, Xiao Li, Bhiksha Raj, and Yan Lu.
\newblock Robust referring video object segmentation with cyclic structural consensus.
\newblock In \emph{Proceedings of the IEEE/CVF International Conference on Computer Vision}, pp.\  22236--22245, 2023{\natexlab{b}}.

\bibitem[Li et~al.(2022{\natexlab{b}})Li, Mao, Girshick, and He]{li2022exploring}
Yanghao Li, Hanzi Mao, Ross Girshick, and Kaiming He.
\newblock Exploring plain vision transformer backbones for object detection.
\newblock In \emph{European Conference on Computer Vision}, pp.\  280--296. Springer, 2022{\natexlab{b}}.

\bibitem[Lin et~al.(2017)Lin, Goyal, Girshick, He, and Doll{\'a}r]{lin2017focal}
Tsung-Yi Lin, Priya Goyal, Ross Girshick, Kaiming He, and Piotr Doll{\'a}r.
\newblock Focal loss for dense object detection.
\newblock In \emph{Proceedings of the IEEE international conference on computer vision}, pp.\  2980--2988, 2017.

\bibitem[Liu et~al.(2023)Liu, Zhu, Li, Chen, Wang, and Shen]{liu2023matcher}
Yang Liu, Muzhi Zhu, Hengtao Li, Hao Chen, Xinlong Wang, and Chunhua Shen.
\newblock Matcher: Segment anything with one shot using all-purpose feature matching.
\newblock \emph{arXiv preprint arXiv:2305.13310}, 2023.

\bibitem[Liu et~al.(2020)Liu, Zhang, Zhang, and He]{liu2020part}
Yongfei Liu, Xiangyi Zhang, Songyang Zhang, and Xuming He.
\newblock Part-aware prototype network for few-shot semantic segmentation.
\newblock In \emph{Computer Vision--ECCV 2020: 16th European Conference, Glasgow, UK, August 23--28, 2020, Proceedings, Part IX 16}, pp.\  142--158. Springer, 2020.

\bibitem[Loshchilov \& Hutter(2017)Loshchilov and Hutter]{loshchilov2017decoupled}
Ilya Loshchilov and Frank Hutter.
\newblock Decoupled weight decay regularization.
\newblock \emph{arXiv preprint arXiv:1711.05101}, 2017.

\bibitem[Ma et~al.(2024{\natexlab{a}})Ma, He, Li, Han, You, and Wang]{ma2024segment}
Jun Ma, Yuting He, Feifei Li, Lin Han, Chenyu You, and Bo~Wang.
\newblock Segment anything in medical images.
\newblock \emph{Nature Communications}, 15\penalty0 (1):\penalty0 654, 2024{\natexlab{a}}.

\bibitem[Ma et~al.(2024{\natexlab{b}})Ma, Kim, Li, Baharoon, Asakereh, Lyu, and Wang]{ma2024segment2}
Jun Ma, Sumin Kim, Feifei Li, Mohammed Baharoon, Reza Asakereh, Hongwei Lyu, and Bo~Wang.
\newblock Segment anything in medical images and videos: Benchmark and deployment.
\newblock \emph{arXiv preprint arXiv:2408.03322}, 2024{\natexlab{b}}.

\bibitem[Ma{\v{s}}ka et~al.(2014)Ma{\v{s}}ka, Ulman, Svoboda, Matula, Matula, Ederra, Urbiola, Espa{\~n}a, Venkatesan, Balak, et~al.]{mavska2014benchmark}
Martin Ma{\v{s}}ka, Vladim{\'\i}r Ulman, David Svoboda, Pavel Matula, Petr Matula, Cristina Ederra, Ainhoa Urbiola, Tom{\'a}s Espa{\~n}a, Subramanian Venkatesan, Deepak~MW Balak, et~al.
\newblock A benchmark for comparison of cell tracking algorithms.
\newblock \emph{Bioinformatics}, 30\penalty0 (11):\penalty0 1609--1617, 2014.

\bibitem[Matsoukas et~al.(2022)Matsoukas, Haslum, Sorkhei, S{\"o}derberg, and Smith]{matsoukas2022makes}
Christos Matsoukas, Johan~Fredin Haslum, Moein Sorkhei, Magnus S{\"o}derberg, and Kevin Smith.
\newblock What makes transfer learning work for medical images: Feature reuse \& other factors.
\newblock In \emph{Proceedings of the IEEE/CVF Conference on Computer Vision and Pattern Recognition}, pp.\  9225--9234, 2022.

\bibitem[Mazurowski et~al.(2023)Mazurowski, Dong, Gu, Yang, Konz, and Zhang]{mazurowski2023segment}
Maciej~A Mazurowski, Haoyu Dong, Hanxue Gu, Jichen Yang, Nicholas Konz, and Yixin Zhang.
\newblock Segment anything model for medical image analysis: an experimental study.
\newblock \emph{Medical Image Analysis}, 89:\penalty0 102918, 2023.

\bibitem[Men{\'e}ndez et~al.(1997)Men{\'e}ndez, Pardo, Pardo, and Pardo]{menendez1997jensen}
Mar{\'\i}a~Luisa Men{\'e}ndez, JA~Pardo, L~Pardo, and MC~Pardo.
\newblock The jensen-shannon divergence.
\newblock \emph{Journal of the Franklin Institute}, 334\penalty0 (2):\penalty0 307--318, 1997.

\bibitem[Milletari et~al.(2016)Milletari, Navab, and Ahmadi]{milletari2016v}
Fausto Milletari, Nassir Navab, and Seyed-Ahmad Ahmadi.
\newblock V-net: Fully convolutional neural networks for volumetric medical image segmentation.
\newblock In \emph{2016 fourth international conference on 3D vision (3DV)}, pp.\  565--571. Ieee, 2016.

\bibitem[Nguyen \& Todorovic(2019)Nguyen and Todorovic]{nguyen2019feature}
Khoi Nguyen and Sinisa Todorovic.
\newblock Feature weighting and boosting for few-shot segmentation.
\newblock In \emph{Proceedings of the IEEE/CVF International Conference on Computer Vision}, pp.\  622--631, 2019.

\bibitem[Oquab et~al.(2023)Oquab, Darcet, Moutakanni, Vo, Szafraniec, Khalidov, Fernandez, Haziza, Massa, El-Nouby, et~al.]{oquab2023dinov2}
Maxime Oquab, Timoth{\'e}e Darcet, Th{\'e}o Moutakanni, Huy Vo, Marc Szafraniec, Vasil Khalidov, Pierre Fernandez, Daniel Haziza, Francisco Massa, Alaaeldin El-Nouby, et~al.
\newblock Dinov2: Learning robust visual features without supervision.
\newblock \emph{arXiv preprint arXiv:2304.07193}, 2023.

\bibitem[Radford et~al.(2018)Radford, Narasimhan, Salimans, Sutskever, et~al.]{radford2018improving}
Alec Radford, Karthik Narasimhan, Tim Salimans, Ilya Sutskever, et~al.
\newblock Improving language understanding by generative pre-training.
\newblock 2018.

\bibitem[Radford et~al.(2019)Radford, Wu, Child, Luan, Amodei, Sutskever, et~al.]{radford2019language}
Alec Radford, Jeffrey Wu, Rewon Child, David Luan, Dario Amodei, Ilya Sutskever, et~al.
\newblock Language models are unsupervised multitask learners.
\newblock \emph{OpenAI blog}, 1\penalty0 (8):\penalty0 9, 2019.

\bibitem[Rakelly et~al.(2018)Rakelly, Shelhamer, Darrell, Efros, and Levine]{rakelly2018conditional}
Kate Rakelly, Evan Shelhamer, Trevor Darrell, Alyosha Efros, and Sergey Levine.
\newblock Conditional networks for few-shot semantic segmentation.
\newblock 2018.

\bibitem[Ravi et~al.(2024)Ravi, Gabeur, Hu, Hu, Ryali, Ma, Khedr, R{\"a}dle, Rolland, Gustafson, et~al.]{ravi2024sam}
Nikhila Ravi, Valentin Gabeur, Yuan-Ting Hu, Ronghang Hu, Chaitanya Ryali, Tengyu Ma, Haitham Khedr, Roman R{\"a}dle, Chloe Rolland, Laura Gustafson, et~al.
\newblock Sam 2: Segment anything in images and videos.
\newblock \emph{arXiv preprint arXiv:2408.00714}, 2024.

\bibitem[Ronneberger et~al.(2015)Ronneberger, Fischer, and Brox]{ronneberger2015u}
Olaf Ronneberger, Philipp Fischer, and Thomas Brox.
\newblock U-net: Convolutional networks for biomedical image segmentation.
\newblock In \emph{Medical Image Computing and Computer-Assisted Intervention--MICCAI 2015: 18th International Conference, Munich, Germany, October 5-9, 2015, Proceedings, Part III 18}, pp.\  234--241. Springer, 2015.

\bibitem[R{\"u}schendorf(1985)]{ruschendorf1985wasserstein}
Ludger R{\"u}schendorf.
\newblock The wasserstein distance and approximation theorems.
\newblock \emph{Probability Theory and Related Fields}, 70\penalty0 (1):\penalty0 117--129, 1985.

\bibitem[Sanderson \& Matuszewski(2022)Sanderson and Matuszewski]{sanderson2022fcn}
Edward Sanderson and Bogdan~J Matuszewski.
\newblock Fcn-transformer feature fusion for polyp segmentation.
\newblock In \emph{Annual conference on medical image understanding and analysis}, pp.\  892--907. Springer, 2022.

\bibitem[Sonke et~al.(2019)Sonke, Aznar, and Rasch]{sonke2019adaptive}
Jan-Jakob Sonke, Marianne Aznar, and Coen Rasch.
\newblock Adaptive radiotherapy for anatomical changes.
\newblock In \emph{Seminars in radiation oncology}, volume~29, pp.\  245--257. Elsevier, 2019.

\bibitem[Strudel et~al.(2021)Strudel, Garcia, Laptev, and Schmid]{strudel2021segmenter}
Robin Strudel, Ricardo Garcia, Ivan Laptev, and Cordelia Schmid.
\newblock Segmenter: Transformer for semantic segmentation.
\newblock In \emph{Proceedings of the IEEE/CVF international conference on computer vision}, pp.\  7262--7272, 2021.

\bibitem[Touvron et~al.(2023)Touvron, Lavril, Izacard, Martinet, Lachaux, Lacroix, Rozi{\`e}re, Goyal, Hambro, Azhar, et~al.]{touvron2023llama}
Hugo Touvron, Thibaut Lavril, Gautier Izacard, Xavier Martinet, Marie-Anne Lachaux, Timoth{\'e}e Lacroix, Baptiste Rozi{\`e}re, Naman Goyal, Eric Hambro, Faisal Azhar, et~al.
\newblock Llama: Open and efficient foundation language models.
\newblock \emph{arXiv preprint arXiv:2302.13971}, 2023.

\bibitem[Vaswani(2017)]{vaswani2017attention}
A~Vaswani.
\newblock Attention is all you need.
\newblock \emph{Advances in Neural Information Processing Systems}, 2017.

\bibitem[Wang et~al.(2019{\natexlab{a}})Wang, Tyagi, Rimner, Hu, Veeraraghavan, Li, Hunt, Mageras, and Zhang]{wang2019segmenting}
Chuang Wang, Neelam Tyagi, Andreas Rimner, Yu-Chi Hu, Harini Veeraraghavan, Guang Li, Margie Hunt, Gig Mageras, and Pengpeng Zhang.
\newblock Segmenting lung tumors on longitudinal imaging studies via a patient-specific adaptive convolutional neural network.
\newblock \emph{Radiotherapy and Oncology}, 131:\penalty0 101--107, 2019{\natexlab{a}}.

\bibitem[Wang et~al.(2020)Wang, Alam, Zhang, Hu, Nadeem, Tyagi, Rimner, Lu, Thor, and Zhang]{wang2020predicting}
Chuang Wang, Sadegh~R Alam, Siyuan Zhang, Yu-Chi Hu, Saad Nadeem, Neelam Tyagi, Andreas Rimner, Wei Lu, Maria Thor, and Pengpeng Zhang.
\newblock Predicting spatial esophageal changes in a multimodal longitudinal imaging study via a convolutional recurrent neural network.
\newblock \emph{Physics in Medicine \& Biology}, 65\penalty0 (23):\penalty0 235027, 2020.

\bibitem[Wang et~al.(2019{\natexlab{b}})Wang, Liew, Zou, Zhou, and Feng]{wang2019panet}
Kaixin Wang, Jun~Hao Liew, Yingtian Zou, Daquan Zhou, and Jiashi Feng.
\newblock Panet: Few-shot image semantic segmentation with prototype alignment.
\newblock In \emph{proceedings of the IEEE/CVF international conference on computer vision}, pp.\  9197--9206, 2019{\natexlab{b}}.

\bibitem[Wang et~al.(2023{\natexlab{a}})Wang, Wang, Cao, Shen, and Huang]{wang2023images}
Xinlong Wang, Wen Wang, Yue Cao, Chunhua Shen, and Tiejun Huang.
\newblock Images speak in images: A generalist painter for in-context visual learning.
\newblock In \emph{Proceedings of the IEEE/CVF Conference on Computer Vision and Pattern Recognition}, pp.\  6830--6839, 2023{\natexlab{a}}.

\bibitem[Wang et~al.(2023{\natexlab{b}})Wang, Zhang, Cao, Wang, Shen, and Huang]{wang2023seggpt}
Xinlong Wang, Xiaosong Zhang, Yue Cao, Wen Wang, Chunhua Shen, and Tiejun Huang.
\newblock Seggpt: Segmenting everything in context.
\newblock \emph{arXiv preprint arXiv:2304.03284}, 2023{\natexlab{b}}.

\bibitem[Wong et~al.(2023)Wong, Rakic, Guttag, and Dalca]{wong2023scribbleprompt}
Hallee~E Wong, Marianne Rakic, John Guttag, and Adrian~V Dalca.
\newblock Scribbleprompt: Fast and flexible interactive segmentation for any medical image.
\newblock \emph{arXiv preprint arXiv:2312.07381}, 2023.

\bibitem[Wu \& Xu(2024)Wu and Xu]{wu2024one}
Junde Wu and Min Xu.
\newblock One-prompt to segment all medical images.
\newblock In \emph{Proceedings of the IEEE/CVF Conference on Computer Vision and Pattern Recognition}, pp.\  11302--11312, 2024.

\bibitem[Wu et~al.(2023)Wu, Fu, Fang, Liu, Wang, Xu, Jin, and Arbel]{wu2023medical}
Junde Wu, Rao Fu, Huihui Fang, Yuanpei Liu, Zhaowei Wang, Yanwu Xu, Yueming Jin, and Tal Arbel.
\newblock Medical sam adapter: Adapting segment anything model for medical image segmentation.
\newblock \emph{arXiv preprint arXiv:2304.12620}, 2023.

\bibitem[Yan et~al.(2023)Yan, Jiang, Wu, Wang, Luo, Yuan, and Lu]{yan2023universal}
Bin Yan, Yi~Jiang, Jiannan Wu, Dong Wang, Ping Luo, Zehuan Yuan, and Huchuan Lu.
\newblock Universal instance perception as object discovery and retrieval.
\newblock In \emph{Proceedings of the IEEE/CVF Conference on Computer Vision and Pattern Recognition}, pp.\  15325--15336, 2023.

\bibitem[Yang et~al.(2021)Yang, Wei, and Yang]{yang2021associating}
Zongxin Yang, Yunchao Wei, and Yi~Yang.
\newblock Associating objects with transformers for video object segmentation.
\newblock \emph{Advances in Neural Information Processing Systems}, 34:\penalty0 2491--2502, 2021.

\bibitem[Zhang et~al.(2025)Zhang, Gao, Jiao, Liu, and Wei]{zhang2025bridge}
Anqi Zhang, Guangyu Gao, Jianbo Jiao, Chi Liu, and Yunchao Wei.
\newblock Bridge the points: Graph-based few-shot segment anything semantically.
\newblock \emph{Advances in Neural Information Processing Systems}, 37:\penalty0 33232--33261, 2025.

\bibitem[Zhang et~al.(2023)Zhang, Jiang, Guo, Yan, Pan, Dong, Gao, and Li]{zhang2023personalize}
Renrui Zhang, Zhengkai Jiang, Ziyu Guo, Shilin Yan, Junting Pan, Hao Dong, Peng Gao, and Hongsheng Li.
\newblock Personalize segment anything model with one shot.
\newblock \emph{arXiv preprint arXiv:2305.03048}, 2023.

\bibitem[Zhang \& Shen(2024)Zhang and Shen]{zhang2024unleashing}
Yichi Zhang and Zhenrong Shen.
\newblock Unleashing the potential of sam2 for biomedical images and videos: A survey.
\newblock \emph{arXiv preprint arXiv:2408.12889}, 2024.

\bibitem[Zou et~al.(2024)Zou, Yang, Zhang, Li, Li, Wang, Wang, Gao, and Lee]{zou2024segment}
Xueyan Zou, Jianwei Yang, Hao Zhang, Feng Li, Linjie Li, Jianfeng Wang, Lijuan Wang, Jianfeng Gao, and Yong~Jae Lee.
\newblock Segment everything everywhere all at once.
\newblock \emph{Advances in Neural Information Processing Systems}, 36, 2024.

\end{thebibliography}
\bibliographystyle{tmlr}
\clearpage

\appendix
\section*{Appendix}
\label{appendix}
\begin{itemize}
    \item \ref{appendix: sam overview}: SAM Overview
    \item \ref{appendix: sam adaptation details}: SAM Adaptation Details
    \item \ref{appendix: test implementation details}: Test Implementation Details
    \item \ref{appendix: additional visualizations}: Additional Visualizations
    \item \ref{appendix: discussions}: Discussions
    \item \ref{appendix: equations}: Equations
\end{itemize}

\section{SAM Overview}
\label{appendix: sam overview}

Segment Anything Model~(SAM)~\citep{kirillov2023segment} comprises three main components: an image encoder, a prompt encoder, and a mask decoder, denoted as $\mathit{Enc}_{I}$, $\mathit{Enc}_{P}$, and $\mathit{Dec}_{M}$. As a promptable segmentation model, SAM takes an image $I$ and a set of human-given prompts $P$ as input. SAM predicts segmentation masks $\mathit{Ms}$ by:
\begin{equation}
    \mathit{Ms} = \mathit{Dec}_{M}(\mathit{Enc}_{I}(I), \mathit{Enc}_{P}(P))
\label{equ: sam}
\end{equation}
During training, SAM supervises the mask prediction with a linear combination of focal loss~\citep{lin2017focal} and dice loss~\citep{milletari2016v} in a 20:1 ratio. When only a single prompt is provided, SAM generates multiple predicted masks. However, SAM backpropagates from the predicted mask with the lowest loss. Note that SAM returns only one predicted mask when presented with multiple prompts simultaneously.

$\mathit{Enc}_{I}$ and $\mathit{Dec}_{M}$ primarily employ the Transformer~\citep{vaswani2017attention, dosovitskiy2020image} architecture. 
Here, we provide details on components in $\mathit{Enc}_{P}$. 
$\mathit{Enc}_{P}$ supports three prompt modalities as input: the point, box, and mask logit. 
The positive- and negative-point prompts are represented by two learnable embeddings, denoted as $E_{\texttt{pos}}$ and $E_{\texttt{neg}}$, respectively. 
The box prompt comprises two learnable embeddings representing the left-up and right-down corners of the box, denoted as $E_{\texttt{up}}$ and $E_{\texttt{down}}$. 
In cases where neither the point nor box prompt is provided, another learnable embedding $E_{\texttt{not-a-point}}$ is utilized. 
If available, the mask prompt is encoded by a stack of convolution layers, denoted as $E_{\texttt{mask}}$; otherwise, it is represented by a learnable embedding $E_{\texttt{not-a-mask}}$. 

SAM employs an interactive training strategy. 
In the first iteration, either a positive-point prompt, represented by $E_{\texttt{pos}}$, or a box prompt, represented by $\{E_{\texttt{up}}, E_{\texttt{down}}\}$, is randomly selected with equal probability from the ground truth mask. 
Since there is no mask prompt in the first iteration, $E_{\texttt{pos}}$ or $\{E_{\texttt{up}}, E_{\texttt{down}}\}$ is combined with $E_{\texttt{not-a-mask}}$ and fed into $\mathit{Dec}_{M}$.
In the follow-up iterations, subsequent positive- and negative-point prompts are uniformly selected from the error region between the predicted mask and the ground truth mask. 
SAM additionally provides the mask logit prediction from the previous iteration as a supplement prompt.
As a result, $\{E_{\texttt{pos}}, E_{\texttt{neg}}, E_{\texttt{mask}}\}$ is fed into $\mathit{Dec}_{M}$ during each iteration.
There are 11 total iterations: one sampled initial input prompt, 8 iteratively sampled points, and two iterations where only the mask prediction from the previous iteration is supplied to the model.

\section{SAM Adaptation Details}
\label{appendix: sam adaptation details}
In \Secref{subsec: adapt sam to medical image domain if needed}, we propose to adapt SAM to the medical image domain when it is needed, with full fine-tune~(\emph{Full-Fine-Tune}) and LoRA~\citep{hu2021lora}~(\emph{LoRA}). 
For \emph{Full-Fine-Tune}, we fine-tune all parameters in SAM backbone.
For \emph{LoRA}, we insert the LoRA module in the image encoder $\mathit{Enc}_{I}$ and only fine-tune parameters in the LoRA module and the mask decoder $\mathit{Dec}_{M}$.
Our fine-tuning objectives are as follows:
\begin{enumerate}
    \item 
    The model can accurately predict a mask even if no prompt is provided.
    \item 
    The model can predict an exact mask even if only one prompt is given.
    \item
    The model maintains promptable ability.
\end{enumerate}
The training strategy outlined in SAM cannot satisfy all these three requirements: 1. The mask decoder $\mathit{Dec}_{M}$ is not trained to handle scenarios where no prompt is given. 2. The approach to resolving the ambiguous prompt by generating multiple results is redundant as we have a well-defined segmentation objective. Despite that, we find a simple modification can meet all our needs:
\begin{enumerate}
    \item 
    In the initial iteration, we introduce a scenario where no prompt is provided to SAM. As a result, $\{E_{\texttt{not-a-point}}, E_{\texttt{not-a-mask}}\}$ is fed into $\mathit{Dec}_{M}$ in the first iteration.
    \item 
    To prevent $E_{\texttt{not-a-point}}$ and $E_{\texttt{not-a-mask}}$ from introducing noise when human-given prompts are available, we stop their gradients in every iteration.
    \item 
    We ensure that SAM always returns an exact predicted mask. As a result, the ambiguity property does not exist in the model after fine-tuning.
\end{enumerate}

\section{Test Implementation Details}
\label{appendix: test implementation details}

\begin{table}[h]
\begin{center}
\vspace{-10pt}
\begin{minipage}[h]{\textwidth}
    \centering
    \caption{Retrieval range for the COCO-$20^i$, FSS-1000, LVIS-$92^i$, PerSeg dataset. {\color{cyan}{Blue}} indicates the retrieval range for positive-point prompts. {\color{red}{Red}} indicates the retrieval range for negative-point prompts.}
    \vspace{5pt}
    \setlength{\tabcolsep}{4pt}
    \begin{tabular}{c c c c}
        \toprule
        COCO-$20^i$ & FSS-1000 & LVIS-$92^i$ & PerSeg\\
        \hline \vspace{-5pt}\\
        $\color{cyan}{1, 5\texttt{-}10}$ / $\color{red}{1}$ & $\color{cyan}{1\texttt{-}5}$ / $\color{red}{1}$ & $\color{cyan}{1, 5\texttt{-}10}$ / $\color{red}{1}$ & $\color{cyan}{1\texttt{-}5}$ / $\color{red}{1}$ \\
        \bottomrule
    \end{tabular}
    \label{tab: appendix_1}
    \vspace{-20pt}
\end{minipage}%
\end{center}
\end{table}
\begin{table}[h]
\begin{center}
\begin{minipage}[h]{\textwidth}
    \centering
    \caption{Retrieval range for the 4D-Lung and CVC-ClinicDB dataset. {\color{cyan}{Blue}} indicates the retrieval range for positive-point prompts. {\color{red}{Red}} indicates the retrieval range for negative-point prompts.}
    \vspace{5pt}
    \setlength{\tabcolsep}{4pt}
    \begin{tabular}{c c c c c c}
        \toprule
        \multirow{2}*{Dataset} & \emph{Meta} & \multicolumn{2}{c}{\emph{LoRA}} & \multicolumn{2}{c}{\emph{Full-Fine-Tune}}\\
        \cmidrule(lr){2-2}\cmidrule(lr){3-4}\cmidrule(lr){5-6}
         \quad & $huge$ & $base$ & $large$ & $base$ & $large$\\
        \hline \vspace{-5pt}\\
        4D-Lung & $\color{cyan}{1\texttt{-}2}$ / $\color{red}{45}$ & $\color{cyan}{1\texttt{-}3}$ / $\color{red}{1}$ & $\color{cyan}{1\texttt{-}3}$ / $\color{red}{1}$ & $\color{cyan}{1\texttt{-}3}$ / $\color{red}{1}$ & $\color{cyan}{1\texttt{-}3}$ / $\color{red}{1}$ \\
        CVC-ClinicDB & $\color{cyan}{1\texttt{-}5}$ / $\color{red}{1\texttt{-}3}$ & $\color{cyan}{1\texttt{-}3}$ / $\color{red}{1\texttt{-}3}$ & $\color{cyan}{1\texttt{-}2}$ / $\color{red}{1\texttt{-}3}$ & $\color{cyan}{1\texttt{-}2}$ / $\color{red}{1}$ & $\color{cyan}{1\texttt{-}5}$ / $\color{red}{1\texttt{-}3}$ \\
        \bottomrule
    \end{tabular}
    \label{tab: appendix_2}
    \vspace{-5pt}
\end{minipage}%
\end{center}
\end{table}

In this section, for reproducibility, we provide the details of the retrieval range during the test time for the COCO-$20^i$~\citep{nguyen2019feature}, FSS-1000~\citep{li2020fss}, LVIS-$92^i$~\citep{liu2023matcher}, and Perseg~\citep{zhang2023personalize} dataset in Table~\ref{tab: appendix_1}, the 4D-Lung~\citep{hugo2016data} and CVC-ClinicDB~\citep{bernal2015wm} dataset in Table~\ref{tab: appendix_2}.

The final number of positive-point and negative-point prompts is determined by our distribution-guided retrieval approach. Below, we explain how the retrieval range is determined in Table~\ref{tab: appendix_2}.
For \emph{LoRA} and \emph{Full-Fine-Tune}, the retrieval range is determined based on the validation set of the \emph{internal} datasets. We uniformly sample positive-point and negative-point prompts on the ground-truth mask and perform interactive segmentation. The number of prompts is increased until the improvement becomes marginal, at which point this maximum number is set as the retrieval range for \emph{external} test datasets.
On the 4D-Lung dataset, we consistently set the number of negative-point prompts to 1 for these two types of models. This decision is informed by conclusions from previous works~\citep{ma2024segment, huang2024segment}, which suggest that the background and semantic target can appear very similar in CT images, and using too many negative-point prompts may confuse the model.
On the CVC-ClinicDB dataset, the endoscopy video is in RGB space, resulting in a relatively small domain gap~\citep{matsoukas2022makes} compared to SAM's pre-trained dataset. Therefore, for \emph{Meta}, we use the same retrieval range as the \emph{Full-Fine-Tune} large model.
In contrast, on the 4D-Lung dataset, CT images are in grayscale, leading to a significant domain gap~\citep{matsoukas2022makes} compared to SAM's pre-trained dataset. Consequently, we set the retrieval range for positive-point prompts to 2 to avoid outliers and fixed the number of negative-point prompts to a large constant (\ie, 45) rather than a range, to ensure the model focuses on the semantic target. These values were not further tuned.

\section{Additional Visualization}
\label{appendix: additional visualizations}

\begin{figure*}[h]
  \centering
  \includegraphics[width=\linewidth]{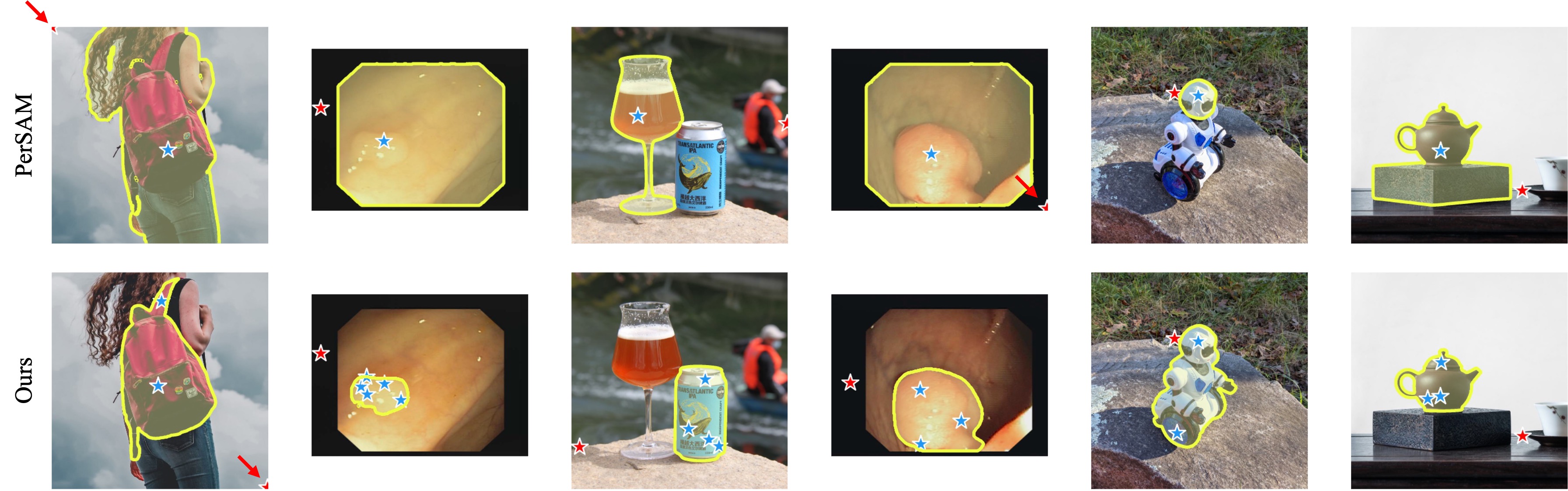}
  \caption{Additional qualitative results: (Columns 1–4) Full images from earlier illustrations; (Columns 5–6) Additional comparisons with PerSAM. Note that the negative-point prompt can sometimes differ between \ppsam and PerSAM, as the similarity matrix changes when using part-level features.}
  \label{fig: appendix_1}
  \vspace{-10pt}
\end{figure*}

\begin{figure*}[h]
    \noindent
    \begin{minipage}[h]{.48\textwidth}
        \centering
        \includegraphics[width=\linewidth]{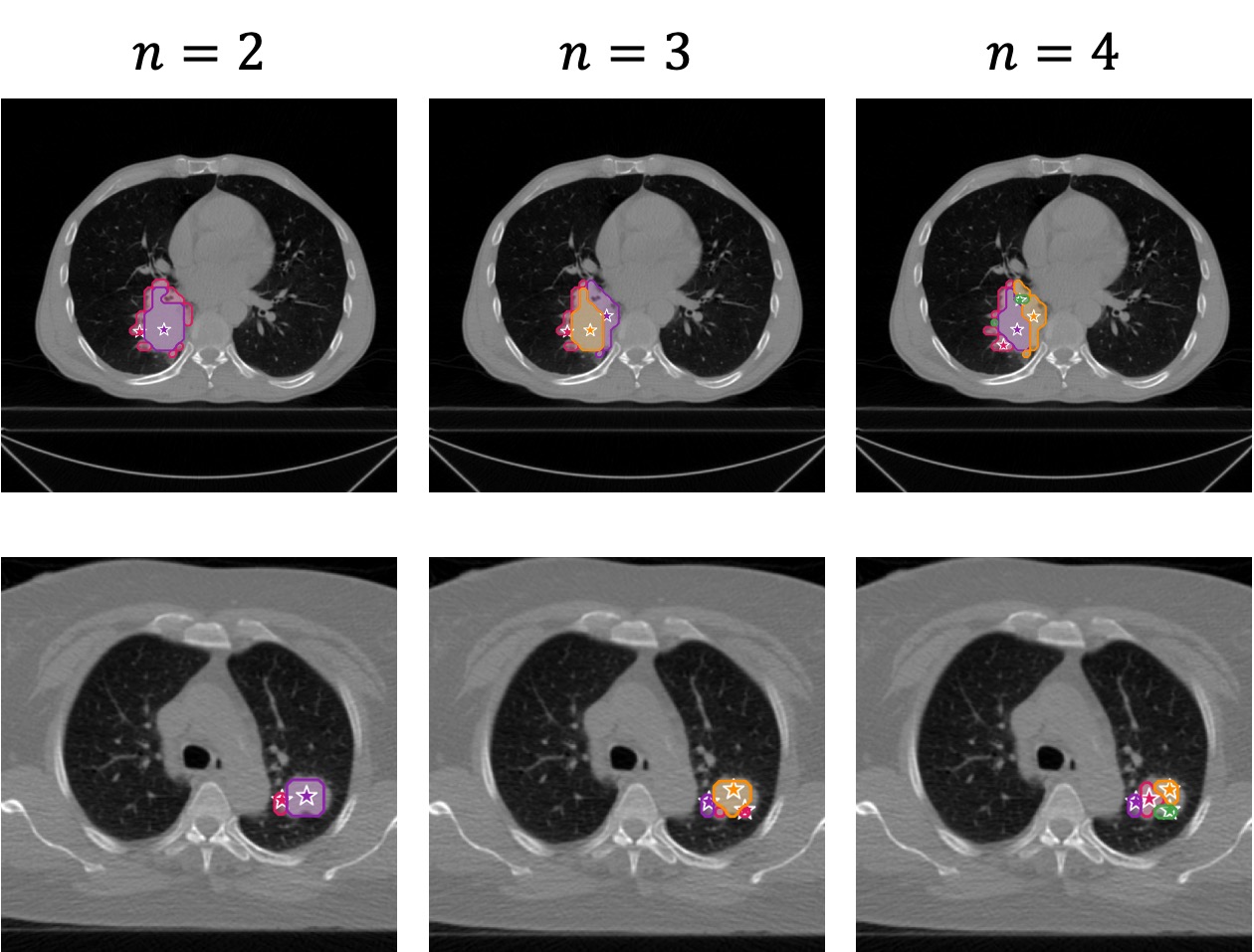} 
        \caption{Visualization results on the 4D-Lung dataset, based on a varying number of part-level features.} 
        \label{fig: appendix_2}
    \end{minipage}%
    \hfill
    \begin{minipage}[h]{.48\textwidth} 
        \centering
        \includegraphics[width=\linewidth]{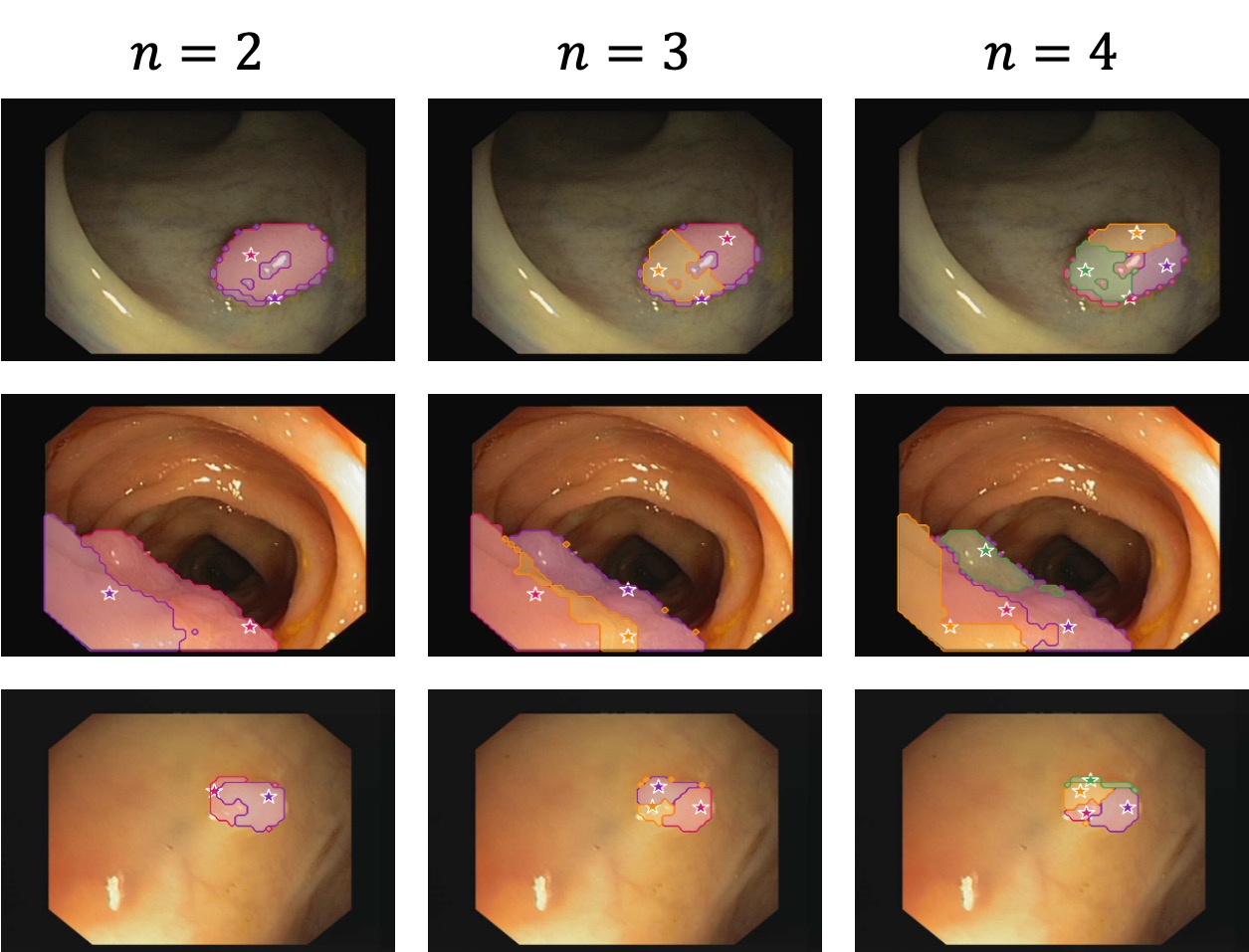}
        \caption{Visualization results on the CVC-ClinicDB dataset, based on a varying number of part-level features.}
        \label{fig: appendix_3}
    \end{minipage}
\end{figure*}

\begin{figure*}[h]
  \centering
  \includegraphics[width=\linewidth]{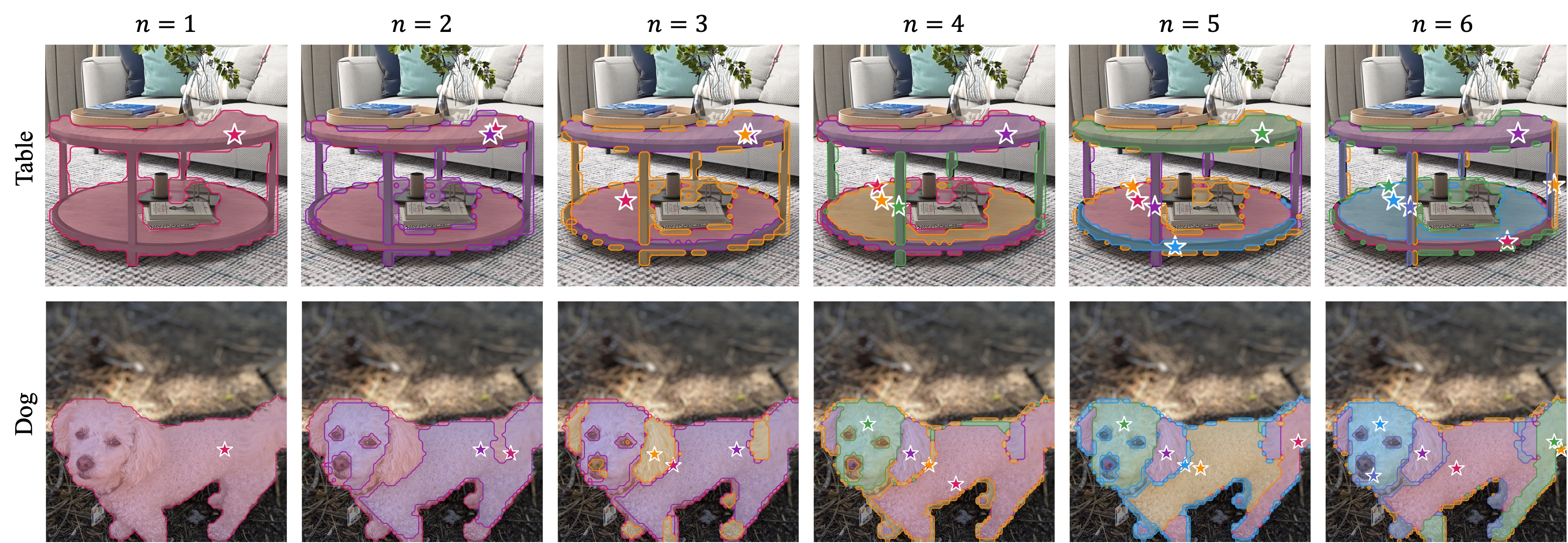}
  \caption{Visualization results on the PerSeg dataset, based on a varying number of part-level features.}
  \label{fig: appendix_4}
\end{figure*}

In this section, we first provide the full images in \Figref{fig: appendix_1} that were presented in \Secref{sec: introduction} to eliminate any possible confusion. Then, to provide deeper insight into our part-aware prompt mechanism and distribution-guided retrieval approach, we present additional visualization results on the 4D-Lung~\citep{hugo2016data} dataset, the CVC-ClinicDB~\citep{bernal2015wm} dataset, and the PerSeg~\citep{zhang2023personalize} dataset. These visualizations are based on a varying number of part-level features, offering a clearer understanding of how the part-aware prompt mechanism adapts to different segmentation tasks and domains.
In Figure~\ref{fig: appendix_2} and~\ref{fig: appendix_3}, we observe that an appropriate number of part-level features can effectively divide the tumor into distinct parts, such as the body and edges for non-small cell lung cancer, and the body and light point~(caused by the camera) for the polyp. This illustrates how \ppsam can assist in cases of incomplete segmentation.
In Figure~\ref{fig: appendix_4}, we observe that an appropriate number of part-level features can effectively divide the object into meaningful components, such as the pictures, characters, and aluminum material of a can; the legs and platforms of a table; or the face, ears, and body of a dog. These parts can merge naturally based on texture features when using the appropriate number of part-level features, whereas using too many features may result in over-segmentation.
Our retrieval approach, on the other hand, helps determine the optimal number of part-level features for each specific case.

\section{Discussion}
\label{appendix: discussions}

\begin{table}[!h]
\noindent
\begin{minipage}[t]{0.42\textwidth}
    \caption{Results of \emph{direct-transfer} on CVC-ClinicDB. The model is trained on Kvasir-SEG with different pre-training weights.}
    \vspace{15pt}
    \begin{center}
    \setlength{\tabcolsep}{2pt}
    \small
    \begin{tabular}{c c c}
        \toprule
        Med-SAM & & 83.85 \\
        \hline \vspace{-10pt} \\
        SAM & & 84.62 \\
        \bottomrule
    \end{tabular}
    \label{tab: appendix_3}
    \end{center}
\end{minipage}%
\hfill
\begin{minipage}[t]{0.54\textwidth}
    \caption{Results of interactive segmentation on the internal Kvasir-SEG validation dataset and the external CVC-ClinicDB dataset. We use \emph{Full-Fine-Tune large$^{\texttt{312.5M}}$ here.}
    }
    \vspace{5pt}
    \begin{center}
    \setlength{\tabcolsep}{2pt}
    \small
    \begin{tabular}{c c c c c}
        \toprule
        Dataset 
        & No Prompt
        & \ppsam
        & 1 Positive
        & Box
        \\
        & & Point Prompt & Prompt \\
        \hline \vspace{-5pt}\\
        Kvasir-SEG
        & 93.27
        & -
        & 95.15
        & 95.57
        \\
        CVC-ClinicDB
        & 86.68
        & 88.76
        & 88.99
        & 92.35
        \\
        \bottomrule
    \end{tabular}
    \label{tab: appendix_4}
    \end{center}
\end{minipage}%
\\
\begin{minipage}[t]{0.42\textwidth}
    \caption{Comparison with GF-SAM on the CVC-ClinicDB dataset. $\star$ indicates using DINOv2 for a better performance.}
    \vspace{10pt}
    \begin{center}
    \setlength{\tabcolsep}{2pt}
    \small
    \begin{tabular}{c c c}
        \toprule
        Method & \emph{Meta} & \emph{Full-Fine-Tune} \\
        \hline \vspace{-5pt} \\
        GF-SAM & 60.55$^\star$ & 87.57 \\
        Matcher & 63.54 & 87.15 \\
        \ppsam & 66.45 & 88.76 \\
        \bottomrule
    \end{tabular}
    \label{tab: appendix_5}
    \end{center}
\end{minipage}%
\hfill
\begin{minipage}[t]{0.54\textwidth}
    \caption{Results of one-shot part segmentation on the PASCAL-Part dataset. 
    Note that all methods utilize SAM's encoder for fairness.
    }
    \vspace{5pt}
    \begin{center}
    \setlength{\tabcolsep}{2pt}
    \small
    \begin{tabular}{c c c c c c}
        \toprule
        Method 
        & animals
        & indoor
        & person
        & vehicles
        & mean
        \\
        \hline \vspace{-5pt}\\
        Matcher
        & 29.29
        & 56.30
        & 21.04
        & 37.02
        & 33.66
        \\
        PerSAM
        & 19.9
        & 51.8 
        & 18.6
        & 32.0
        & 30.1
        \\
        \ppsam
        & 20.29
        & 54.82
        & 19.62
        & 34.21
        & 32.24
        \\
        \ppsam \emph{w. neg}
        & 20.54
        & 54.65
        & 20.59
        & 36.91
        & 33.17
        \\
        \bottomrule
    \end{tabular}
    \label{tab: appendix_6}
    \end{center}
\end{minipage}
\end{table}

\noindent\textbf{Baseline Results.}
In this paper, we treat MedSAM~\citep{ma2024segment} with a human-given box prompt as the baseline for the 4D-Lung dataset~\citep{hugo2016data}, DuckNet~\citep{dumitru2023using} as the baseline for the CVC-ClinicDB dataset~\citep{bernal2015wm}.
We acknowledge that MedSAM is widely used as a baseline across many benchmarks~\citep{antonelli2022medical, ji2022amos}. However, these comparisons primarily focus on internal validation. MedSAM has the potential to outperform many models on external validation sets due to its pre-training on a large-scale medical image dataset. While there is no direct evidence to confirm this, DuckNet~\citep{dumitru2023using} suggests that large-scale pre-trained models generally outperform others on external validation sets, even if they lag behind on internal validation.
Among studies~\citep{butoi2023universeg, wong2023scribbleprompt, ma2024segment, ma2024segment2, wu2024one} that aim to develop promptable segmentation models specifically for medical image segmentation, UniverSeg~\citep{butoi2023universeg}'s performance may decline significantly with only one-shot support set, and both ScribblePrompt~\citep{wong2023scribbleprompt} and One-Prompt~\citep{wu2024one} are trained on much smaller datasets. 
As we focus on segmenting external patient samples that lie outside the training distribution in a one-shot manner. Therefore, we argue that the model's generalization ability is critical for achieving superior performance. 
The 4D-Lung dataset~\citep{hugo2016data} is a relatively new benchmark for longitudinal data analysis, and no standard benchmark for comparison was available at the time this work was conducted.
In addition, during evaluation, we supplemented MedSAM with a human-given box prompt, making it a very fair baseline for this work.

\noindent\textbf{Baseline Methods.} 
In this paper, we treat SAM-based methods such as PerSAM~\citep{zhang2023personalize} and Matcher~\citep{liu2023matcher} as our primary baselines and also compare with PANet~\citep{wang2019panet}. We do not include other backbone methods like ScribblePrompt~\citep{wong2023scribbleprompt} and One-Prompt~\citep{wu2024one} because they primarily focus on interactive segmentation, just similar to MedSAM~\citep{ma2024segment}, which is the baseline we compare in \Tabref{tab: result3}.
On the other hand, utilizing other prompt modalities, such as scribble, mask, and box,
presents challenges for solving the patient-adaptive segmentation problem, as it is difficult to represent prior data in these formats. In this work, we adopt a more flexible prompt modality: point prompts. Although it may be possible to convert our multiple-point prompts into a scribble prompt by connecting them together, we leave the exploration of this direction for future work. Consequently, the most relevant baseline methods remain SAM-base methods like PerSAM and Matcher. 

Here, we evaluate a more recent SAM-base method, GF-SAM~\citep{zhang2025bridge}. Similar to Matcher, GF-SAM utilizes DINOv2 to extract patch-level features; however, GF-SAM is a hyper-parameter-free method based on graph analysis. In Table~\ref{tab: appendix_5}, we evaluate GF-SAM on the CVC-ClinicDB dataset~\citep{bernal2015wm} using both a natural image pre-trained encoder~(\emph{Meta}) and a medically adapted encoder~(\emph{Full-Fine-Tune}). With the natural image pre-trained encoder, \ppsam outperforms both GF-SAM and Matcher, since patch-level features are less robust than part-level features when there is domain gap between pre-training data and test data. However, GF-SAM fails to surpass Matcher in this task, which contrasts with its superior performance on natural image segmentation tasks. We hypothesize that this is because GF-SAM is a hyper-parameter-free method, and factors such as the number of point prompts, the number of clusters, and the threshold value may be more sensitive when there is a domain gap between the pre-training data and the test data. GF-SAM outperforms Matcher with the medically adapted encoder, but still lags behind \ppsam, as the encoder is adapted for medical segmentation tasks and still lacks patch-level objectives. This result, along with the findings in Table~\ref{tab: ablation3}—where \ppsam with \emph{base} SAM outperforms PerSAM with \emph{huge} SAM by 0.7\% mIoU and \emph{base} SAM by 26.0\% mIoU on the PerSeg dataset—further underscores that \ppsam is a more robust method when the model exhibits weaker representations, a scenario more prevalent in medical image analysis.

\noindent\textbf{Pre-trained Model.} In this work, we choose to adapt SAM to the medical image domain using the SA-1B pre-trained model weights rather than weights from MedSAM for two reasons. First, although MedSAM fine-tunes SAM~(SA-1B pre-trained) on a large-scale medical segmentation dataset, its fine-tuning dataset is still 1,000 times smaller than SAM’s pre-training dataset (1M vs. 1B). Since model generality after adaptation is crucial for our work, we assume that SAM remains a better starting point, despite MedSAM being a strong option for zero-shot medical segmentation. Second, MedSAM only provides the SAM-Base pre-trained model, whereas our results in Table~\ref{tab: result1} and Table~\ref{tab: result2} demonstrate that larger models (i.e., \emph{large}) can further enhance performance across various tasks. In Table~\ref{tab: appendix_3}, we provide the \emph{direct-transfer} result on the CVC-ClinicDB dataset, the model is trained on the Kvasir-SEG dataset with Med-SAM pre-trained weights and SA-1B pre-trained weights. The result follows our assumption and the discussion in MedSAM~\citep{ma2024segment} and its successor~\citep{ma2024segment2}, that with a specific task, maybe fine-tune from SAM is still a better choice.

\noindent\textbf{Interactive Segmentation.} As mentioned in \Secref{subsec: adapt sam to medical image domain if needed} and detailed in Appendix~\ref{appendix: sam adaptation details}, we closely adhere to SAM's interactive training strategy when adapting it with medical datasets. Therefore, our medically adapted model retains its interactive segmentation capability. In Table~\ref{tab: appendix_4}, we present both internal evaluation results on the Kvasir-SEG dataset's validation set and external evaluation results on the CVC-ClinicDB dataset.
First, as discussed in Section~\ref{subsec: quantitative results} and Appendix~\ref{appendix: sam adaptation details}, since we have a specific segmentation target, our adapted model does not need to be ambiguity-aware, allowing a human-given single positive-point prompt to achieve good performance.
\ppsam lags only slightly behind this result while operating fully automatically. For the human-given box prompt, it is not surprising that it outperforms \ppsam, as a box prompt is a strong prompt that essentially requires the provider to know the lesion's location.

\noindent\textbf{Part Segmentation.}
We acknowledge that P2SAM's design was not initially focused on part segmentation but on enhancing the medical image segmentation model's generality by providing more precise and informative prompts. 
We conduct the part segmentation task on the PASCAL-Part dataset~\citep{everingham2010pascal}. Note that all methods use SAM~(\emph{Meta}) as the backbone model. Part segmentation with SAM typically relies more on additional prompt modalities, such as box prompts, or diverse mask candidates. For example, Matcher employs a random point-prompt sampling strategy to make their proposed mask candidates more diverse, potentially slowing down the algorithm. 
In Table~\ref{tab: appendix_6}, when compared with PerSAM, \ppsam consistently shows benefits (i.e., +2.23\% mIoU). However, \ppsam is surpassed by Matcher~(i.e., -1.42\% mIoU). 
For \ppsam it is reasonable to provide additional negative-point prompts in part segmentation task because a portion of the background is correlated between the reference and target images (i.e., both refer to the rest of the object). Therefore, we additionally provide negative-point prompts to \ppsam~(\ppsam \emph{w.} neg), which further improves segmentation performance (i.e., +0.93\% mIoU) and brings \ppsam on par with Matcher. While achieve slightly better performance, Matcher utilizes 128 sampling iterations for the part segmentation task, making it much slower (x3) than both PerSAM and \ppsam. 

\begin{figure*}[t]
  \centering
  \includegraphics[width=0.9 \linewidth]{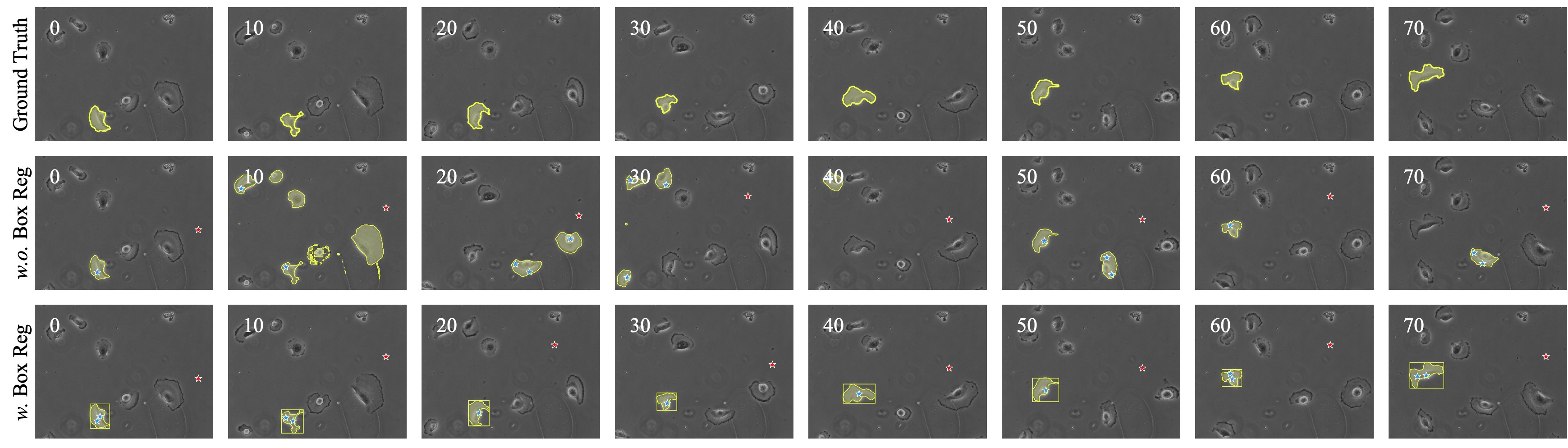}
  \caption{Qualitative results of single-cell segmentation on the PhC-C2DH-U373 dataset. The second row highlights the challenge \ppsam faces in handling multiple similar objects. The third row demonstrates that \ppsam can overcome this challenge with a cost-free regularization.}
  \label{fig: appendix_5}
\end{figure*}

\noindent\textbf{Similar Objects.}
\ppsam demonstrates improvements in the backbone's generalization across domain, task, and model levels. At the task level, we have already shown how \ppsam enhances performance for NSCLC segmentation in patient-adaptive radiation therapy and polyp segmentation in endoscopy videos. However, when addressing specific tasks that involve multiple similar targets, \ppsam may fail due to the lack of instance-level objective. Although this scenario is uncommon in patient-adaptive segmentation, we acknowledge that \ppsam faces the same challenge of handling multiple similar objects as other methods~\citep{zhang2023personalize, liu2023matcher}. In Figure~\ref{fig: appendix_5}, we present an example of single-cell segmentation on the PhC-C2DH-U373 dataset~\citep{mavska2014benchmark}, which goes beyond the patient-specific setting. 
In Figure~\ref{fig: appendix_5}, the second row illustrates that \ppsam fails to segment the target cell due to the presence of many similar cells in the field of view. However, given the slow movement of the cell, we can leverage its previous information to regularize the current part-aware prompt mechanism. 
The third row in Figure~\ref{fig: appendix_5} demonstrates that when using the bounding box from the last frame, originally propagated from the reference frame, to regularize the part-aware prompt mechanism in the current frame, \ppsam achieves strong performance on the same task. Since the bounding box for the first frame can be generated from the ground truth mask, which is already available, this regularization incurs no additional cost. Utilizing such tailored regularization incorporating various prompt modalities, we showcase our approach’s flexible applicability to other applications.

\section{Equations}
\label{appendix: equations}
In this section, we provide details on the equation mentioned in Section~\ref{subsec: methodology overview}. 

\noindent\textbf{Wasserstein Distance.} In Equation~\ref{equ: distribution-guided retrieval}, we use $\mathcal{D}_{w}(\cdot, \cdot)$ to represent the Wasserstein Distance. Here we provide the details of this function. Suppose that features in the reference image $F_R \in \mathbb{R}^{n_r \times d}$ and features in the target image $F_T \in \mathbb{R}^{n_t \times d}$ come from two discrete distributions, $F_R \in \mathbf{P}(\mathbb{F}_{\mathbb{R}})$ and $F_T \in \mathbf{P}(\mathbb{F}_\mathbb{R})$, where $F_R = \sum_{i=1}^{n_r} u_i \delta^{i}_{f_r}$ and $F_T = \sum_{j=1}^{n_t} v_j \delta^{j}_{f_t}$; $\delta_{f_r}$ being the Delta-Dirac function centered on $f_r$ and $\delta_{f_t}$ being the Delta-Dirac function centered on $f_t$. Since $F_R$ and $F_T$ are both probability distributions, sum of weight vectors is $1$, $\sum_{i}u_{i} = 1 = \sum_{j}v_j$. The Wasserstein distance between $F_R$ and $F_T$ is defined as: 
\begin{equation}\label{equ: wasserstein distance}
\mathcal{D}_{w}(F_R,F_T) = \min_{\mathbf{T}\in \Pi(u,v)} \sum_{i} \sum_{j} \mathbf{T}_{ij} \cdot  \frac{F^{i}_{R} \cdot F^{j}_{T}}{{\left \| F^{i}_{R} \right \|}_{2} \cdot {\left \| F^{j}_{T} \right \|}_{2}}
\end{equation}

where $\Pi(u,v) = \{ \mathbf{T} \in \mathbb{R}_+^{n\times m} | \mathbf{T}\mathbf{1}_m=u, \mathbf{T}^\top\mathbf{1}_n=v \}$, and $\mathbf{T}$ is the transport plan, interpreting the amount of mass shifted from $F^i_R$ to $F^j_T$.

\end{document}